\documentclass{article}
\usepackage{geometry}
\usepackage[T1]{fontenc}
\usepackage{lmodern}
\usepackage[utf8]{inputenc}
\usepackage{tgtermes}

\usepackage{hyperref}
\usepackage{caption}
\usepackage{subcaption}
\usepackage{amsmath,amssymb,amsfonts}
\usepackage{algorithm}
\usepackage{algpseudocode}
\usepackage{graphicx}
\usepackage{textcomp}
\usepackage{float}
\usepackage[numbers,sort]{natbib}
\usepackage{booktabs}

\allowdisplaybreaks

\usepackage[shortlabels]{enumitem}

\usepackage{titlesec}
\titlespacing*{\section}
{0pt}{1.1ex plus .2ex minus .2ex}{0.8ex plus .2ex minus .2ex}
\titlespacing*{\subsection}
{0pt}{0.8ex plus .2ex minus .2ex}{0.5ex plus .2ex minus .2ex}

\usepackage{array}
\newcommand{\PreserveBackslash}[1]{\let\temp=\\#1\let\\=\temp}
\newcolumntype{C}[1]{>{\PreserveBackslash\centering}p{#1}}
\newcolumntype{R}[1]{>{\PreserveBackslash\raggedleft}p{#1}}
\newcolumntype{L}[1]{>{\PreserveBackslash\raggedright}p{#1}}

\newcommand{\Identity}{{\rm I\kern-.2em l}}

\usepackage{amsthm}
\newtheorem{definition}{Definition}

\newtheorem{theorem}{Theorem}

\usepackage{microtype}
\usepackage{graphicx}

\usepackage{booktabs} 
\usepackage{array}
 \usepackage{amssymb}

\usepackage[utf8]{inputenc} 
\usepackage[T1]{fontenc}   
\usepackage{url}            
\usepackage{booktabs}       
\usepackage{amsfonts,apxproof}       
\usepackage{nicefrac}       
\usepackage{microtype}      
\usepackage{xcolor}         
\usepackage{xspace}
\usepackage{wrapfig}
\usepackage{multirow}

\usepackage{amsmath}
\usepackage{amssymb}
\usepackage{mathtools}
\usepackage{amsthm}
\usepackage{amsfonts,comment,bbm}
\usepackage[capitalize,noabbrev]{cleveref}
\usepackage{xspace}

\theoremstyle{plain}
\theoremstyle{remark}
\newtheorem{remark}[theorem]{Remark}

\usepackage[textsize=tiny]{todonotes}

\newcommand{\bef}{\begin{figure}}
\newcommand{\eef}{\end{figure}}
\newcommand{\beq}{\begin{eqnarray}}
\newcommand{\eeq}{\end{eqnarray}}

\newcommand{\DFC}{\texttt{Cached-DFL}\xspace}

\begin{document}

\title{Decentralized Federated Learning with Model Caching on Mobile Agents}

\author{
Xiaoyu Wang$^{1}$, Guojun Xiong$^{2}$, Houwei Cao$^3$, Jian Li$^2$, Yong Liu$^1$\\
$^1$New York University, $^2$Stony Brook University, $^3$New York Institute of Technology\\ \{wang.xiaoyu, yongliu\}@nyu.edu, \{guojun.xiong, jian.li.3\}@stonybrook.edu, \\hcao02@nyit.edu
\date{}
}

\maketitle
\makeatletter
\def\blfootnote{\xdef\@thefnmark{}\@footnotetext}
\makeatother

\begin{abstract}
Federated Learning (FL) trains a shared model using data and computation power on distributed agents  coordinated by a central server. Decentralized FL (DFL) utilizes local model exchange and aggregation between agents to reduce the communication and computation overheads on the central server. However, when agents are mobile, the communication opportunity between agents can be sporadic, largely hindering the convergence and accuracy of DFL. In this paper, we propose \textbf{Cached} \textbf{D}ecentralized \textbf{F}ederated \textbf{L}earning (\DFC) to investigate delay-tolerant model spreading and aggregation enabled by model caching on mobile agents. Each agent stores not only its own model, but also models of  agents encountered in the recent past. When two agents meet, they exchange their own models as well as the cached models. Local model aggregation utilizes all models stored in the cache. We theoretically analyze the convergence of \DFC, explicitly taking into account the model staleness introduced by caching. We design and compare different model caching algorithms for different DFL and mobility scenarios. We conduct detailed case studies in a vehicular network to systematically investigate the interplay between agent mobility, cache staleness, and model convergence. In our experiments, \DFC converges quickly, and significantly outperforms DFL without caching.
\end{abstract}

\section{Introduction}\label{intro}

\subsection{Federated Learning on Mobile Agents}
Federated learning (FL) is a type of distributed machine learning (ML) that prioritizes data  privacy~\cite{mcmahan2017communication}. The traditional FL involves a central server that connects with a large number of agents. The agents retain their data and do not share them with the server. During each communication round, the server sends the current global model to the agents, and a small subset of agents are chosen to update the global model by running stochastic gradient descent (SGD)~\cite{robbins1951stochastic} for multiple iterations on their local data. The central server then aggregates the updated parameters to obtain the new global model. 
FL naturally complements emerging Internet-of-Things (IoT) systems, where each IoT device not only can sense its surrounding environment to collect local data, but also is equipped with computation resources for local model training, and communication interfaces to interact with a central server for model aggregation. Many IoT devices are mobile, ranging from mobile phones, autonomous cars/drones, to self-navigating robots. In recent research efforts on smart connected vehicles, there has been a focus on integrating vehicle-to-everything (V2X) networks with Machine Learning (ML) tools and distributed decision making~\cite{barbieri2022decentralized}, particularly in the area of computer vision tasks such as traffic light and signal recognition, road condition sensing, intelligent obstacle avoidance, and intelligent road routing, etc.  
With FL, vehicles locally train deep ML models and upload the model parameters to the central server. 
This approach not only reduces bandwidth consumption, as the size of model parameters is much smaller than  the size of raw image/video data, but also leverages computing power on vehicles, and protects user privacy. 

However, FL on mobile agents still faces communication and computation challenges. The movements of mobile agents, especially at high speed, lead to fast-changing channel conditions on the wireless connections between mobile agents and the central server, resulting in high latency in FL~\cite{niknam2020federated}. Battery-powered mobile agents also have limited power budget for long-range wireless communications. Non-i.i.d data distributions on mobile agents make it difficult for local models to converge. As a result, FL on mobile agents to obtain an optimal global model remains an open challenge. 
Decentralized FL (DFL) has emerged as a potential solution where local model aggregations are conducted between neighboring mobile agents using local device-to-device (D2D) communications with high bandwidth, low latency and low power consumption~\cite{10251949}
. Preliminary studies have demonstrated that DFL algorithms have the potential to significantly reduce the high communication costs associated with centralized FL. However, blindly applying model aggregation algorithms, such as FedAvg~\cite{mcmahan2017communication}, developed for centralized FL to DFL cannot achieve fast convergence and high model accuracy~\cite{liu2022distributed}.

\subsection{Delay-Tolerant Model Communication and Aggregation Through Caching}
D2D model communication between a pair of mobile agents is possible only if they are within each other's transmission ranges. If mobile agents only meet with each others sporadically, there will not be enough model aggregation opportunity for fast convergence. In addition, with non-i.i.d data distributions on agents, if an agent only 
meets with agents from a small cluster, there is no way for the agent to interact with models trained by data samples outside of its cluster, leading to disaggregated local models that cannot perform well on the global data distribution. {It is therefore essential to achieve {\bf fast and even} model spreading using limited D2D communication opportunities among mobile agents.}  A similar problem was studied in the context of Mobile Ad hoc Network (MANET), where wireless communication between mobile nodes are sporadic. The efficiency of data dissemination in MANET can be significantly improved by {\it Delay-Tolerant Networking (DTN)}~\cite{DTN1,burleigh2003delay}: a mobile node caches data it received from nodes it met in the past; when meeting with a new node, it not only transfers its own data, but also the cached data of other nodes. Essentially, node mobility forms a new ``communication" channel through which cached data are transported through node movement in physical space. It is worth noting that, due to multi-hop caching-and-relay, DTN transmission incurs longer delay than D2D direct transmission. Data staleness can be controlled by caching and relay algorithms to match the target application's delay tolerance such as \citet{li2023predictive}.

{\it Motivated by DTN, we propose delay-tolerant DFL communication and aggregation enabled by model caching on mobile agents.} To realize DTN-like model spreading, each mobile agent stores not only its own local model, but also local models received from other agents in the recent history. Whenever it meets another agent, it transfers its own model as well as the cached models to the agent through high-speed D2D communication. Local model aggregation on an agent works on all its cached models, mimicking a local parameter server. Compared with DFL without caching, DTN-like model spreading can push local models faster and more evenly to the whole network;  
aggregating all cached models can facilitate more balanced learning than pairwise model aggregation. While DFL  model caching sounds promising, it also faces a new challenge of {\it model staleness}: a cached model from an agent is not the current model on that agent, with the staleness determined by the mobility patterns, as well as the model spreading and caching algorithms. Using stale models in model aggregation may slow down or even deviate model convergence.

The key challenge we want to address in this paper is how to design cached model spreading and aggregation algorithms 
to achieve fast convergence and high accuracy in DFL on mobile agents. Towards this goal, we make the following contributions: 
\begin{enumerate}
\item We develop \DFC, a new DFL  framework  that utilizes model caching on mobile agents to realize {\it delay-tolerant} model communication and aggregation; 
\item We theoretically analyze the convergence of aggregation with cached models, explicitly taking into account the model staleness; 
\item We design and compare different model caching algorithms for different DFL and mobility scenarios.
\item We conduct a detailed case study on vehicular network to systematically investigate the interplay between agent mobility, cache staleness, and convergence of model  aggregation. Our experimental results demonstrate that our \DFC converges quickly and significantly outperforms DFL without caching.
\end{enumerate}

\begin{table}
\centering
  \caption{Notations and Terminologies.}
  \label{tab:Dataset& Model}
  \begin{tabular}{|c|c|}
    \hline
    \textbf{Notation} & \textbf{Description} \\
\hline
    $N$ & Number of agents\\\hline
    $T$& Number of global epochs\\\hline
    $[N]$& Set of integers $\{1,...,N\}$ \\\hline
    $K$& Number of local updates\\ \hline
    $x_i(t)$& Model in the $t^{th}$ epoch on agent $i$\\ \hline
     $x(t)$& Global Model in the $t^{th}$ epoch,
     $x(t)=\mathbb{E}_{i\in[N]}[x_i(t)]$\\\hline
     $x_i(t,k)$& Model initialized from $x_t$, after  
    $k$-th local update on agent $i$ \\ \hline
    $\Tilde{x}_i(t)$& Model $x_i(t)$ after local updates \\ \hline
    $\mathcal{D}^i$ & Dataset on the $i$-th agent\\ \hline
    $\alpha$& Aggregation weight\\ \hline
    $t-\tau$& Staleness\\ \hline
    $\tau_{max}$& Tolerance of staleness in cache\\ \hline
    $||\cdot||$& All the norms in the paper are $l_2$-norms \\ \hline
\end{tabular}
\end{table}

\section{Background}

\subsection{Global Training Objective}\label{Problem forumation}

Similar to the standard FL problem, the overall objective of mobile DFL is to learn a single global statistical model from data stored on tens to potentially millions of mobile agents. The overall goal is to find the optimal model weights $x^*\in \mathbb{R}^d$ to minimize the global loss function:
\begin{equation}
    \min_x F(x),\, \text{where} \, F(x)=\frac{1}{N}\sum_{i\in [N]}\mathbb{E}_{z^i\sim\mathcal{D}^i}{f(x;z^i)}, \label{FL_Equaltion}
\end{equation}
where $N$ denotes the total number of mobile agents, and each agent has its own local dataset, i.e., $\mathcal{D}^i\neq \mathcal{D}^j,\forall i\neq j$. And $z^i$ is sampled from the local data $\mathcal{D}^i$.

\subsection{DFL Training with Local Model Caching}
\label{Methodology}
All agents participate in DFL training over $T$ global epochs. At the beginning of  the $t^{th}$ epoch, agent $i$'s local model is $x_i(t)$. After $K$ steps of SGD to solve the following optimization problem with a regularized loss function:  
\[\min_x \mathbb{E}_{z^i\sim D^i}f(x;z^i)+\frac{\rho}{2}||x-x_i(t)||^2,\]  
agent $i$ obtains an updated local model $\Tilde{x}_i(t)$. Meanwhile, during the $t^{th}$ epoch, driven by their mobility patterns, each agent meets and exchanges models with other agents. Other than its own model, agent $i$ also stores models it received from other agents encountered in the recent history in its local cache $\mathcal{C}_i(t)$. When two agents meet, they not only exchange their own local models, but also share their cached models with each others to maximize the efficiency of DTN-like model spreading. The models received by agent $i$ will be used to update its model cache $\mathcal{C}_i(t)$, following different cache update algorithms, such as LRU update method (Algorithm \ref{alg: LRU}) or Group-based LRU update method, which will be described in details later. As the cache size of each agent is limited, it is important to design an efficient cache update rule in order to maximize the caching benefit.

After cache updating, each agent conducts local model aggregation using all the cached models with customized aggregation weights $\{\alpha_j \in (0,1)\}$ to get the updated local model $x_i(t+1)$ for epoch $t+1$. In our simulation, we take the aggregation weight as $\alpha_j = (n_j /\sum_{j \in \mathcal{C}_i(t)} n_j )$, where  $n_j $ is the number of samples on agent $j$. 

The whole process repeats until the end of $T$ global epochs. The detailed algorithm is shown in Algorithm \ref{alg: Cached-DFL}. $z^i_k$ are randomly drawn local data samples on agent $i$ for the $k$-th local update, and $\eta$ is the learning rate.

\begin{algorithm}[t]
    \caption{\textbf{Cached} \textbf{D}ecentralized \textbf{F}ederated \textbf{L}earning (\DFC)} \label{alg: Cached-DFL}
    \textbf{Input:} Global epochs $T$, local updates $K$, initial models $\{x_i(0)\}_{i=1}^N$, staleness tolerance $\tau_{\text{max}}$

    \begin{algorithmic}[1]
        \Statex
        \Function{LocalUpdate}{$x_i(t)$}
            \State Initialize: $x_i(t,0) = x_i(t)$
            \State Define: $g_{x(t)}(x;z) = f(x;z) + \frac{\rho}{2}\|x - x(t)\|^2$
            \For{$k = 1, 2, \ldots, K$}
                \State Randomly sample $z_k^i \sim \mathcal{D}^i$
                \State $x_i(t,k) = x_i(t,k-1) - \eta \nabla g_{x_i(t)}(x_i(t,k-1); z_k^i)$
            \EndFor
            \State \Return $\tilde{x}_i(t) = x_i(t,K)$
        \EndFunction

        \Statex
        \Function{ModelAggregation}{$\mathcal{C}_i(t)$}
            \State $x_i(t+1) = \sum_{j \in \mathcal{C}_i(t)} \alpha_j \tilde{x}_j(\tau)$
            \State \Return $x_i(t+1)$
        \EndFunction

        \Statex
        \textbf{Main Process:}
        \For{$t = 0, 1, \ldots, T-1$}
            \For{$i = 1, 2, \ldots, N$}
                \State $\tilde{x}_i(t) \gets$ \Call{LocalUpdate}{$x_i(t)$}
                \State $\mathcal{C}_i(t) \gets$ \Call{CacheUpdate}{$\mathcal{C}_i(t-1), \tau_{\text{max}}$}
                \State $x_i(t+1) \gets$ \Call{ModelAggregation}{$\mathcal{C}_i(t)$}
            \EndFor
        \EndFor
    \end{algorithmic}

    \textbf{Output:} $\{x_i(T)\}_{i=1}^N$
\end{algorithm}

\begin{remark}
Note the $N$ agents communicate with each others in a mobile D2D network. D2D communication can only happen between an agent and its neighbors within a short range (e.g. several hundred meters). Since agent locations are constantly changing (for instance, vehicles continuously move along the road network of a city), D2D network topology is dynamic and can be sparse at any given epoch. To ensure the eventual model convergence, the union graph of D2D networks over multiple epochs should be strongly-connected for efficient DTN-like model spreading.  We also assume D2D communications are non-blocking, carried by short-distance high-throughput communication methods such as mmWave or WiGig, which has enough capacity to complete the exchange of cached models before the agents go out of each other's communication ranges.  
\end{remark}

\begin{remark}
Intuitively, comparing to the DFL without cache (e.g. DeFedAvg~\cite{sun2022decentralized}), where each agent can only get new model by averaging with another model, \DFC uses more models (delayed versions) for aggregation, thus utilizes more underlying information from datasets on more agents. Although \DFC introduces stale models, it can benefit model convergence, especially in highly heterogeneous data distribution scenarios. Overall, our \DFC framework allows each agent to act as a local proxy for Centralized FL with delayed cached models, thus speedup the convergence especially in highly heterogeneous data distribution scenarios, that are challenging for the traditional DFL to converge.
\end{remark}
 
\begin{remark}
    As mentioned above, our approach inevitably introduces stale models. Intuitively, larger staleness results in greater error in the global model. For the cached models with large staleness $t-\tau$, we could set a threshold $\tau_{max}$ to kick old models out of model spreading, which is described in the cache update algorithms. The practical value for $\tau_{max}$ should be related to the cache capacity and communication frequency between agents. In our experimental results, we choose $\tau_{max}$ to be 10 or 20 epochs. Results and analysis about the effect of different $\tau_{max}$ can be found in experimental results section.
\end{remark}

\section{Convergence Analysis}\label{COnvergence analysis}
We now theoretically investigate the impact of caching, especially the staleness of cached models, on DFL model convergence. We introduce some definitions and assumptions. 

\begin{definition}[Smoothness]
A differentiable function $f$  is $L$-smooth if for $\forall x,y,$ $f(y)-f(x)\leq \langle  \nabla f(x),y-x \rangle+\frac{L}{2}||y-x||^2$, where $L>0$.
\end{definition}

\begin{definition}[Bounded Variance]
  There exists constant $\varsigma>0$ such that the global variability of the local gradient of the loss function is bounded $||\nabla F_j(x)-\nabla F(x)||^2\leq \varsigma^2,\forall j \in[N],x\in \mathbb{R}^d$.  
\end{definition}

\begin{definition}[L-Lipschitz Continuous Gradient]
  There exists a constant $L>0$, such that $||\nabla F(x)-\nabla F(x')||\leq L'||x-x'||,\forall x,x'\in \mathbb{R}^d.$  
\end{definition}

\begin{theorem}
Assume that $F$ is $L$-smooth and convex, and each agent executes $K$ local updates before meeting and exchanging models, after that, then does model aggregation. We also assume bounded staleness $\tau < \tau_{max}$, as the kick-out threshold. Furthermore, we assume, $\forall x\in \mathbb{R}^d, i \in [N]$, and $\forall z\sim \mathcal{D}^i, ||\nabla f(x;z)||^2\leq V,||\nabla g_{x'}(x;z)||^2\leq V,\forall x'\in \mathbb{R}^d$, and $\nabla F$ satisfies L-Lipschitz Continuous Gradient. For any small constant $\epsilon>0$, if we take $\rho>0$, and satisfying $-(1+2\rho+\epsilon)V+(\rho^2-\frac{\rho}{2})||x(t,k-1)-x(t)||^2\geq 0,\forall x(t,k-1),x(t)$, after $T$ global epochs, Algorithm \ref{alg: Cached-DFL} converges to a critical point: 
\begin{align}\nonumber
    \min_{t=0}^{T-1} \mathbb{E}||\nabla F(x(t))||^2&\leq\frac{\tau_{max}}{\epsilon\eta C_1KT} \mathbb{E}[F(x(0))-F(x_{M(T)}(T))]+\mathcal{O}(\frac{\eta \rho K^2}{\epsilon C_1})\\
    &\leq \mathcal{O}(\frac{\tau_{max}}{\epsilon\eta C_1KT}) +\mathcal{O}(\frac{\eta \rho K^2}{\epsilon C_1}). 
\end{align}
\end{theorem}

\subsection{Proof Sketch} 
We now highlight the key ideas and challenges behind our convergence proof. 

\noindent{\bf Step 1:} Similar to Theorem 1 in~\citet{DBLP:journals/corr/abs-1903-03934}, we bound the expected cost reduction after $K$ steps of local updates on the 
$j$-th agent, $\forall j\in [N]$, as 
\begin{equation*}
\begin{split}
     \mathbb{E}[F(\Tilde{x}_j(t))-F(x_j(t))] &=\mathbb{E}[F(x_j(t,K))-F(x_j(t,0))]\\
    &\leq -\eta \epsilon \sum_{k=0}^{K-1}\mathbb{E}||\nabla F(x_j(t,k))||^2+\eta^2\mathcal{O}(\rho K^3V).
\end{split} 
\end{equation*}

\noindent{\bf Step 2:} For any epoch $t$, 
we find the index $M(t)$ of the agent whose model is the ``worst", i.e., $M(t) = \arg\max_{j \in [N]}\{F(x_j(t))\}$, and the ``worst" model on all agents over the time period of $[t-\tau_{max}+1, t]$ as 
\[ \mathcal{T}(t,\tau_{max})= \arg\max_{t \in [t-\tau_{max}+1, t]} \{F(x_{M(t)}(t))\}.\]

\noindent{\bf Step 3:} We bound the cost reduction of the ``worst" model at epoch $t+1$ from the ``worst" model in the time period of $[t-\tau_{max}+1, t]$, i.e., the worst possible model that can be stored in some agent's cache at time $t$, as: 
\begin{align}\nonumber\label{eq:bound1}
   \mathbb{E}[F\left(x_{M(t+1)}(t+1)\right)& -F\left(x_{M(\mathcal{T}(t,\tau_{max}))} (\mathcal{T}(t,\tau_{max}))\right)]\\
   &\leq -\epsilon\eta C_1\mathrlap{K} \min_{\tau=t}^{t-\tau_{max}+1} ||\nabla F(x(\tau))||^2+\eta^2\mathcal{O}(\rho K^3V). 
\end{align}
\noindent{\bf Step 4:} We iteratively construct a time sequence  $\{T'_0,T'_1,T'_2,...,T'_{N_T}\}\subseteq \{0,1,...,T-1\}$ in the backward fashion so that 
 \begin{eqnarray*}
    &T'_{N_T} &= T-1;\\
    &T'_i &= \mathcal{T}(T'_{i+1},\tau_{max})-1, \quad 1 \le i \le N_T-1;\\    
    &T'_0 &= 0.
\end{eqnarray*}

\noindent{\bf Step 5:} Applying inequality~(\ref{eq:bound1}) at all time instances $\{T'_0,T'_1,T'_2,...,T'_{N_T}\}$, after T global epochs, we have,
\begin{align}\label{eq:gradientbound}
   \min_{t=0}^{T-1} \mathbb{E}&[||\nabla F(x(t))||^2] \leq \frac{ 1}{N_T}\sum_{t=T'_0}^{T'_{N_T}}\min_{\tau=t}^{t-\tau_{max}+1} ||\nabla F(x(\tau))||^2\nonumber\\
        &\leq \frac{1}{\epsilon\eta C_1KN_T}\sum_{t=T'_0}^{T'_{N_T}} \mathbb{E}[F(x_{M(\mathcal T(t,\tau_{max}))}(\mathcal T(t,\tau_{max})))-F(x_{M(t+1)}(t+1))]+\mathcal{O}(\frac{\eta\rho K^2V}{\epsilon C_1})  \nonumber\\
        &\leq \frac{1}{\epsilon\eta C_1KN_T} \mathbb{E}[F(x(0))-F(x_{M(T)}(T)]+\mathcal{O}(\frac{\eta\rho K^2V}{\epsilon C_1})  \nonumber \\
         &\leq \mathcal{O}(\frac{\tau_{max}}{\epsilon\eta C_1KT}) +\mathcal{O}(\frac{\eta\rho  K^2}{\epsilon C_1}).
\end{align}

\noindent{\bf Step 6:} With the results in~(\ref{eq:gradientbound}) and by leveraging Theorem 1 in \citet{yang2021achieving}, \DFC converges to a critical point after $T$ global epochs.

\section{Realization of \texttt{DFedCache}}
We implement \DFC\footnote{https://github.com/ShawnXiaoyuWang/Cached-DFL} with PyTorch \cite{paszke2017automatic} on Python3.
\subsection{Datasets}
We conduct experiments using three standard FL benchmark datasets: MNIST \cite{deng2012mnist}, FashionMNIST \cite{xiao2017fashionmnistnovelimagedataset}, CIFAR-10 \cite{krizhevsky2009learning} on $N=100$ vehicles, with CNN, CNN and ResNet-18~\cite{he2016deep} as models respectively. 
We evaluate three different data distribution settings: {\it non-i.i.d, i.i.d, and Dirichlet}. In {\it extreme non-i.i.d}, we use a setting similar to \citet{su2022boost}, data points in training set are sorted by labels and then evenly divided into 200 shards, with each shard containing 1--2 labels out of 10 labels. Then 200 shards are randomly assigned to 100 vehicles unevenly: 10\% vehicles receive 4 shards, 20\% vehicles receive 3 shards, 30\% vehicles receive 2 shards and the rest 40\% receive 1 shard. For {\it i.i.d}, we randomly allocate all the training data points to 100 vehicles. For {\it Dirichlet distribution}, we follow the setting in \citet{xiong2024deprl}, to take a heterogeneous allocation by sampling $p_i\sim Dir_N(\pi)$, where $\pi$ is the parameter of Dirichlet distribution. We take $\pi=0.5$ in our following experiments.

\subsection{Evaluation Setup}
The baseline algorithm is DeFedAvg \cite{sun2022decentralized}, which implements simple decentralized federated optimization. For convenience, we name DeFedAvg as DFL in the following results. We set batch size to 64 in all experiments. For MNIST and FashionMNIST, we use 60k data points for training and 10k data points for testing. For CIFAR-10, we use  50k data points for training and 10k data points for testing. Different from training set partition, we do not split the testset. For MNIST and FashionMNIST, we test local models of 100 vehicles on the 10k data points of the whole test set and get the average test accuracy for the evaluation metric. What's more, for CIFAR-10, due to the computing overhead, we sample 1,000 data points from test set for each vehicle and use the average test accuracy of 100 vehicles as the evaluation metric. For all the experiments, we train for 1,000 global epochs, and implement early stop when the average test accuracy stops increasing for at least 20 epochs. For MNIST and FashionMNIST experiments, we use 10 compute nodes, each with 10 CPUs, to simulate DFL on 100 vehicles. CIFAR-10 results are obtained from 1 compute node with 5 CPUs and 1 A100 NVIDIA GPU. 
\begin{algorithm}[t]
    \caption{LRU Model Cache Update (LRU Update)} \label{alg: LRU}
    \textbf{Input:} Current cache $\mathcal{C}_i(t)$, agent $j$'s cache $\mathcal{C}_j(t)$, model $x_j(t)$ from agent $j$, current time $t$, cache size $\mathcal{C}_{\text{max}}$, staleness tolerance $\tau_{\text{max}}$

    \begin{algorithmic}[1]
        \Statex
        \textbf{Main Process:}
        \For{each $x_k(\tau) \in \mathcal{C}_i(t)$ or $\mathcal{C}_j(t)$}\label{line:kick_out}
            \If{$t - \tau \geq \tau_{\text{max}}$}
                \State Remove $x_k(\tau)$ from the respective cache ($\mathcal{C}_i(t)$ or $\mathcal{C}_j(t)$)
            \EndIf
        \EndFor\label{line:end kick_out}

        \State Add or replace $x_j(t)$ into $\mathcal{C}_i(t)$\label{line:Add directly model}

        \For{each $x_k(\tau) \in \mathcal{C}_j(t)$}\label{line:Add cached models}
            \If{$x_k(\tau) \notin \mathcal{C}_i(t)$}
                \State Add $x_k(\tau)$ into $\mathcal{C}_i(t)$
            \Else
                \State Retrieve $x_k(\tau') \in \mathcal{C}_i(t)$
                \If{$\tau > \tau'$}\label{line:LRU metic}
                    \State Replace $x_k(\tau')$ with $x_k(\tau)$ in $\mathcal{C}_i(t)$
                \EndIf
            \EndIf
        \EndFor \label{line:End Add cached models}

        \State Sort models in $\mathcal{C}_i(t)$ in descending order of $\tau$\label{line:sort LRU}
        \State Retain only the first $\mathcal{C}_{\text{max}}$ models in $\mathcal{C}_i(t)$

        \State \Return $\mathcal{C}_i(t+1)$
    \end{algorithmic}

    \textbf{Output:} $\mathcal{C}_i(t+1)$
\end{algorithm}

\subsection{Optimization Method} We use SGD as the optimizer and set the initial learning rate $\eta = 0.1$ for all experiments, and use learning rate scheduler named \textbf{ReduceLROnPlateau} from PyTorch, to automatically adjust the learning rate for each training.

\subsection{Mobile DFL Simulation}
\begin{figure}[t]
\centering
\includegraphics[width=0.7\textwidth]{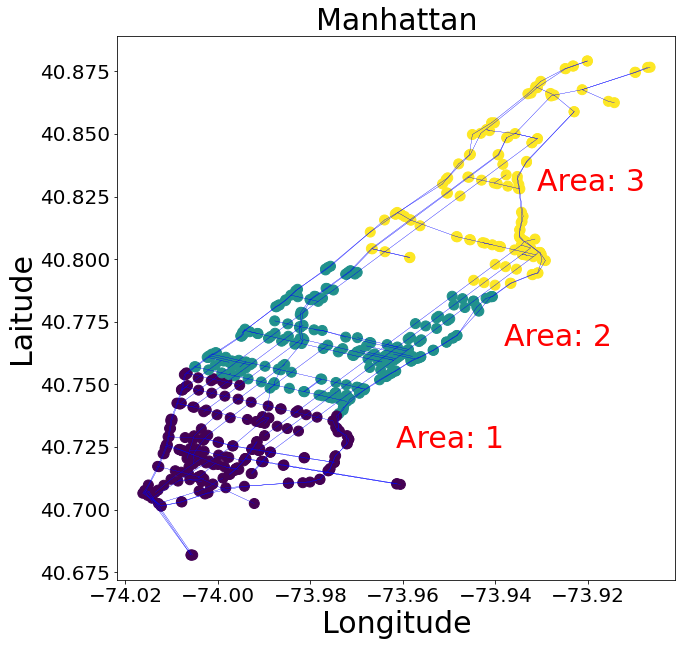}
\caption{\label{fig:3_area} Manhattan Mobility Model Map. The dots represent the intersections while the edges between nodes represent road in Manhattan.}
\end{figure}

Manhattan Mobility Model maps are generated from the real road data of Manhattan from INRIX\footnote{© 2024 INRIX, Inc.,"INRIX documentation," INRIX documentation \url{https://docs.inrix.com/}.}, which is shown in Fig. \ref{fig:3_area}. We follow the setting  from \citet{bai2003important}: a vehicle is allowed to move along the grid of horizontal and vertical streets on the map. At an intersection, the vehicle can turn left, right or go straight. This choice is probabilistic: the probability of moving on the same street is 0.5, and the probabilities of turn into one of the rest roads are equal. For example, at an intersection with 3 more roads to turn into, then the probability to turn into each road is 0.1667. Follow the setting from \citet{su2022boost}: each vehicle is equipped with DSRC and mmWave communication components, can communicate to other vehicles within the communication range of 100 meters, while the default velocity of each vehicle is 13.89 m/s. In our experiments, we set up the number of local updates $K=10$, and the interval between each epoch is $120$ seconds. Within each global epoch, vehicles train their models on local datasets while moving on the road and updating their cache by uploading and downloading models with other encountered  vehicles.
\subsection{LRU Cache Update}
Algorithm \ref{alg: LRU}, which we name as LRU method for convenience, is the basic cache update method we proposed. Basically, the LRU updating rule aims to fetch the most recent models and keep as many of them in the cache as possible. At lines \ref{line:LRU metic} and \ref{line:sort LRU}, the metric for making choice among models is the timestamp of models, which is defined as the epoch time when the model was received from the original vehicle, rather than received from the cache. 
What's more, to fully utilize the caching mechanism, a vehicle not only fetches other vehicles' own trained models, but also models in their caches. For instance, at epoch $t$, vehicle $i$ can directly fetch model $x_j(t)$ from vehicle $j$, at line \ref{line:Add directly model}, and also fetch models $x_k(\tau)\in \mathcal{C}_j(t)$ from the cache of vehicle $j$, at lines \ref{line:Add cached models}-\ref{line:End Add cached models}. This way, each vehicle can not only directly fetch its neighbors' models, but also indirectly fetch models of its neighbors' neighbors, thus boosting the spreading of the underlying data information from different vehicles, and improving the DFL convergence speed, especially with heterogeneous data distribution. Additionally, at lines \ref{line:kick_out}-\ref{line:end kick_out}, before updating cache, models with staleness $t-\tau\geq\tau_{max}$ will be removed from each vehicle's cache.

\begin{figure}[H]
\centering
\includegraphics[width=1\textwidth]{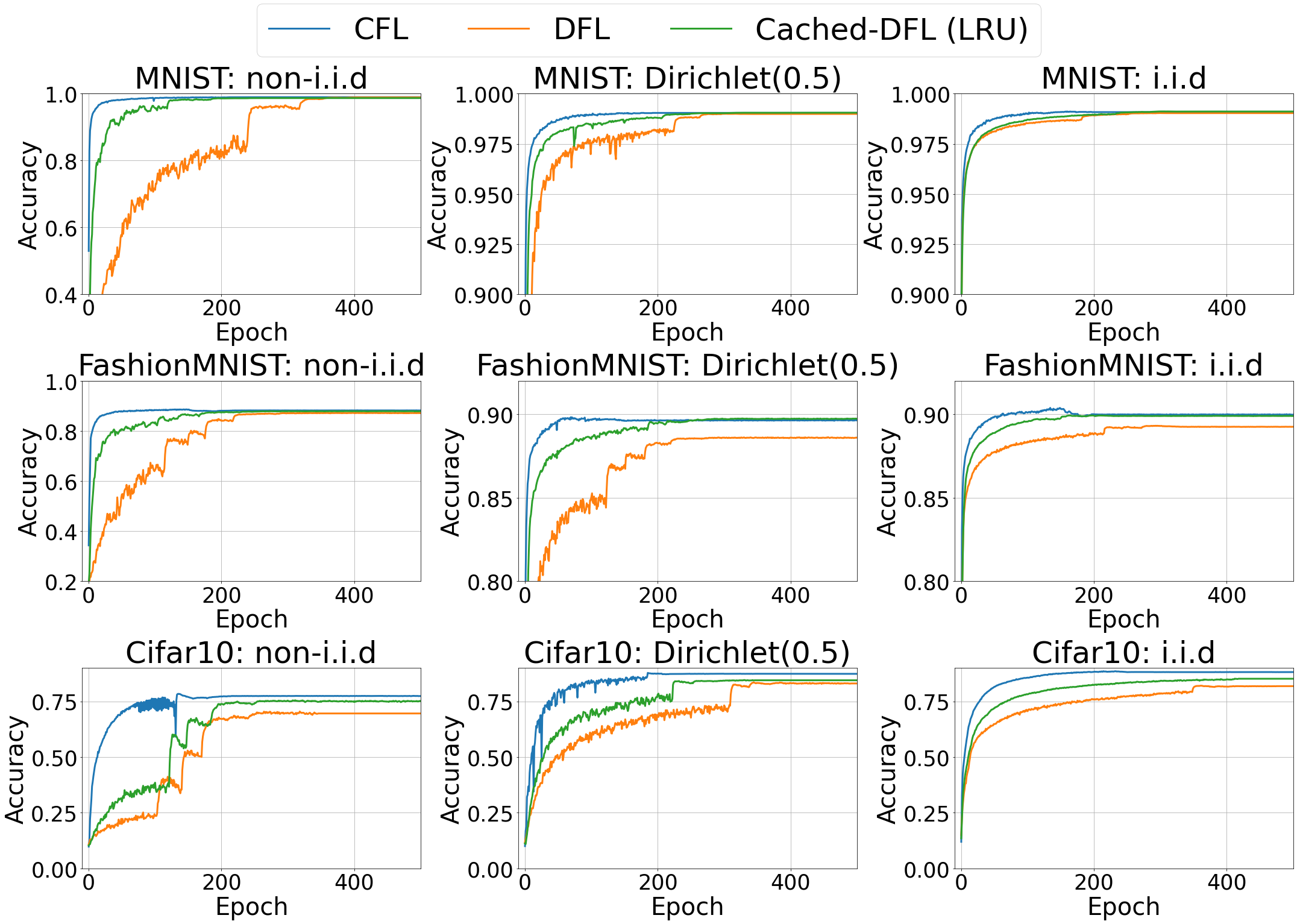}
\caption{\label{fig:cache_non_cache} DFL with Caching vs. DFL without Caching.}
\end{figure}

\section{Numerical Results}\label{sec:sim}

\subsection{Caching vs. Non-caching}
To evaluate the performance of \DFC, we compare the DFL with LRU Cache, Centralized FL (CFL) and DFL on MNSIT, FashionMNIST, CIFAR-10 with three distributions: non-i.i.d, i.i.d, Dirichlet with $\pi=0.5$. For LRU update, we take the cache size as 10 for MNIST and FashionMNIST, and 3 for CIFAR-10 and $\tau_{max}=5$ based on practical considerations. Given a speed of $13.89m/s$ and and a communication distance of $100m$, the communication window of two agents driving in opposite directions could be limited. Additionally, the above cache sizes were chosen by considering the size of chosen models and communication overhead.  From the results in Fig. \ref{fig:cache_non_cache}, we can see that \DFC boosts the convergence and outperforms non-caching method (DFL) and gains performance much closer to CFL in all the cases, especially in non-i.i.d. scenarios.

\subsection{Impact of Cache Size}
\begin{figure}[H]
\centering
\includegraphics[width=1\textwidth]{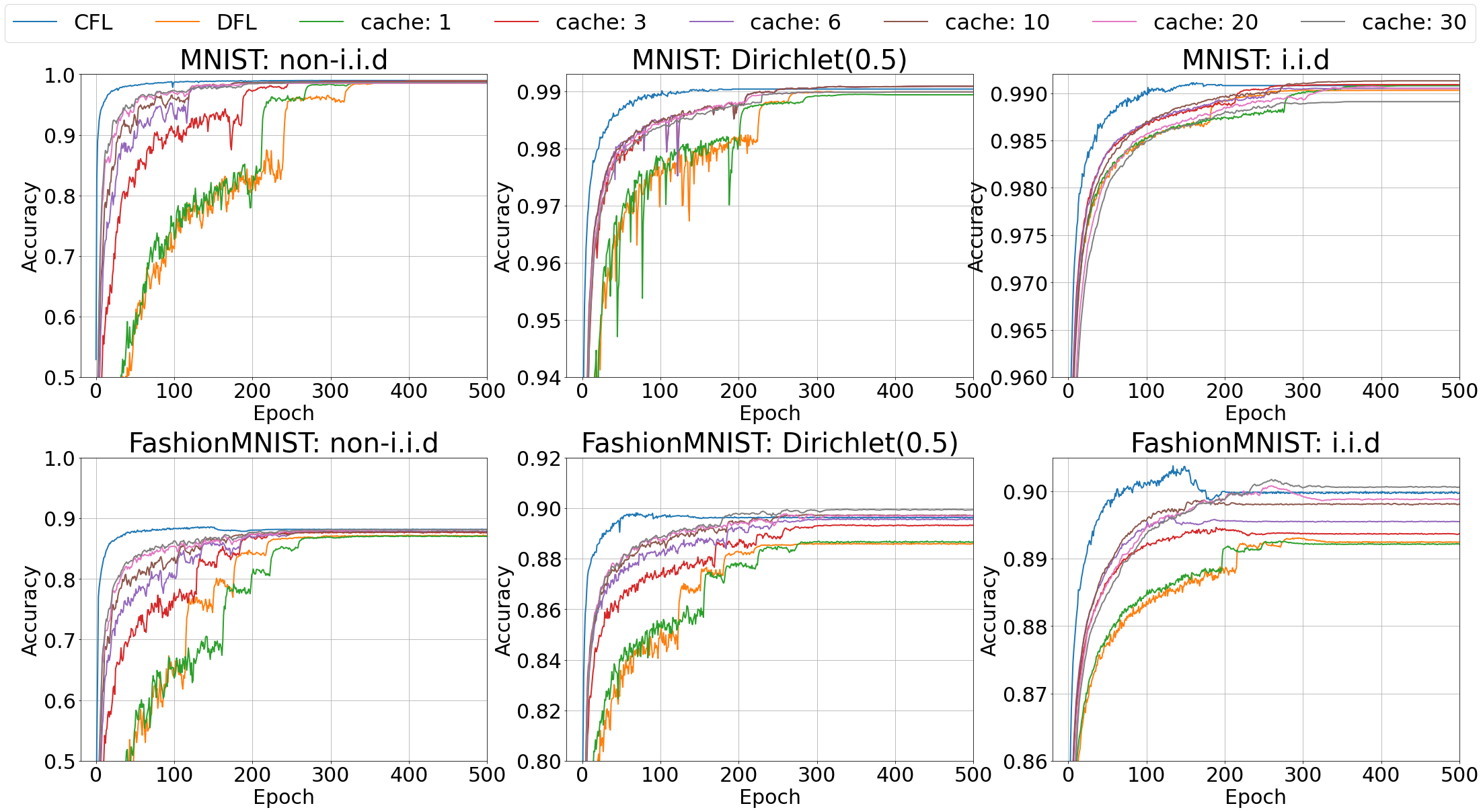}
\caption{\label{fig:different_cache} DFL with LRU at Different Cache Sizes.}
\end{figure}

Then we evaluate the performance gains with different cache sizes from 1 to 30 and  $\tau_{max}=10$, on MNIST and FashionMNIST in Fig. \ref{fig:different_cache}. LRU can benefit more from larger cache sizes, especially in non-i.i.d scenarios, as aggregation with more cached models gets closer to CFL and training with global data distribution.

\subsection{Impact of Model Staleness}
One drawback of model  caching is introducing stale models into aggregation, so it is very important to choose a proper staleness tolerance $\tau_{max}$. First, we statistically calculate the relation between the average number and the average age of cached models at different $\tau_{max}$ from 1 to 20, when epoch time is 30s, 60s, 120s, with unlimited cache size, in Table  \ref{tab:cache_size_age_tradeoff}. We can see that, with the fixed epoch time, as the $\tau_{max}$ increases, the average number and average age of cached models increase, approximately linearly. It's not hard to understand, as every epoch each agent can fetch a limited number of models  directly from other agents. Increasing the staleness tolerance $\tau_{max}$ will increase the number of cached models, as well as  the age of cached models. What's more, we can see that the communication frequency or the moving speed of agents will also impact the average age of cached models, as the faster an agent moves, the more models it can fetch within each epoch, which we will further discuss later.

\begin{table}
\centering
  \caption{Average number and average age of cached models with different $\tau_{max}$ and different epoch time: 30s, 60s, 120s. Columns for different $\tau_{max}$, and rows for different epoch time. There are two sub-rows for each epoch time, with the first sub-row be the average number of cached models, the second sub-row be their average age.}
  \label{tab:cache_size_age_tradeoff}
  \begin{tabular}{|c|r|r|r|r|r|r|r|r|}
    \hline
    &$\tau_{max}$ & 1&2&3&4&5&10&20 \\
\hline
    \multirow{2}*{ 30s} &num &0.8549&1.6841&2.6805&3.8584&5.2054&14.8520&44.8272\\
    \cline{2-9}
    ~&age&0.0000&0.4904&1.0489&1.6385&2.2461&5.4381&11.5735\\ \hline
    \multirow{2}*{ 60s} &num&1.6562&3.8227&6.4792&10.3889&14.0749&43.0518&90.2576\\
    \cline{2-9}
    ~&age&0.0000&0.5593&1.1604&1.8075&2.4396&5.5832&9.7954\\ \hline
    \multirow{2}*{ 120s}  &num&3.6936&9.9149&18.5016&31.3576&35.1958&90.2469&98.5412\\
    \cline{2-9}
    ~&age&0.0000&0.6189&1.2715&1.9194&1.5081&4.7131&5.2357\\ \hline
\end{tabular}
\end{table}

\begin{figure}
\centering
\includegraphics[width=1\textwidth]{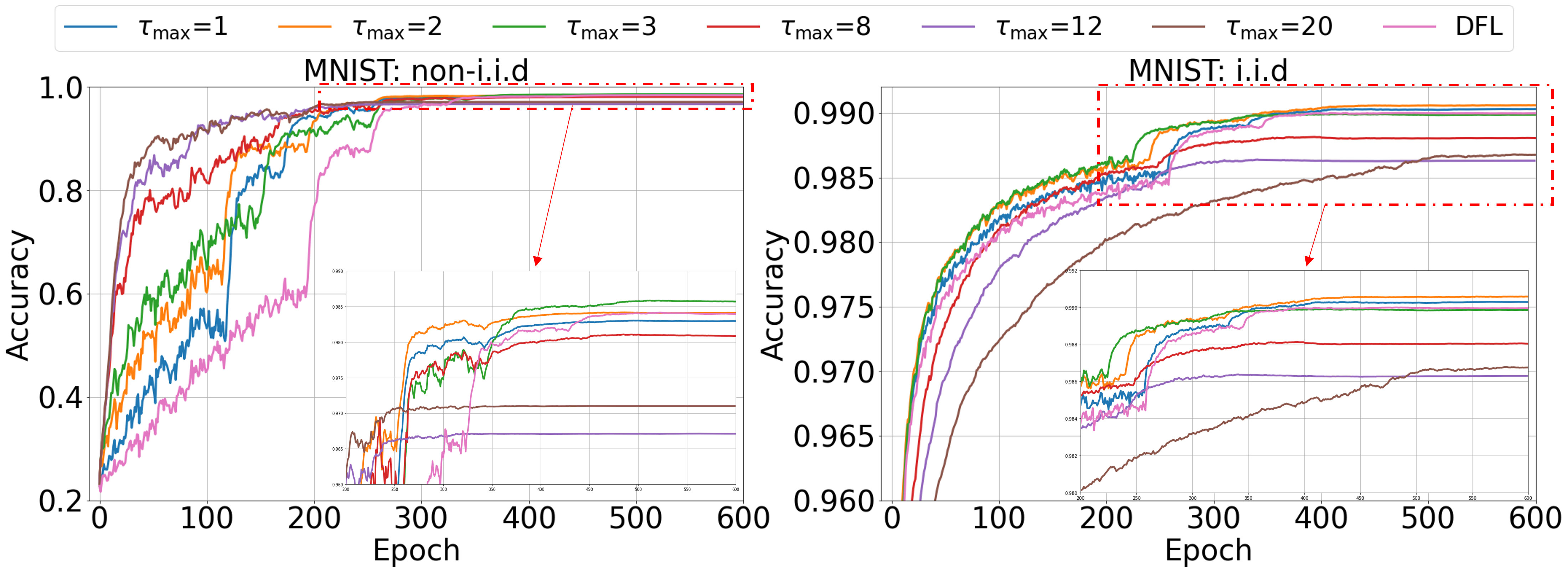}
\caption{\label{fig:different_kick_out} Impact of  $\tau_{max}$ on Model Convergence.}
\end{figure}

We also compare the performance of DFL with LRU at  different $\tau_{max}$ on MNIST  under non-i.i.d and i.i.d in Fig. \ref{fig:different_kick_out}. Here we pick the epoch time 30s for a clear view. First, for non-i.i.d, a larger $\tau_{max}$ brings faster convergence at the beginning,  which is consistent with our previous conclusion, as it allows for more cached models which bring more benefits than harm to the training with non-i.i.d. However, for i.i.d scenarios, larger $\tau_{max}$ stops improving the performance of models, as introducing more cached models hardly benefits training in i.i.d, and the model staleness will prevent convergence. What's more, if we zoom in the final converging phase, which is amplified in the bottom of each figure, we observe that larger $\tau_{max}$ will reduce the final converged accuracy in both non-i.i.d and i.i.d scenarios, as it introduces more staleness. Even with large $\tau_{max}$ and large staleness, LRU method still achieves similar or even better performance than DFL in non-i.i.d. scenarios. 
\subsection{Mobility's Impact on Convergence}
\begin{figure}[H]
\centering
\includegraphics[width=1\textwidth]{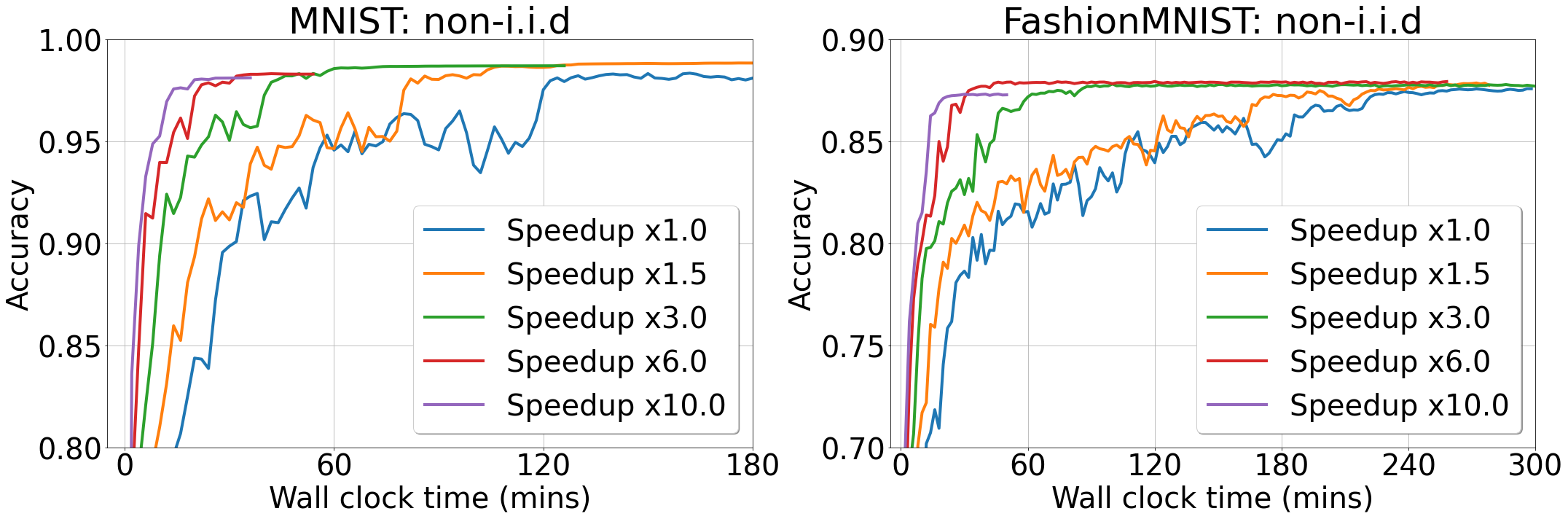}
\caption{\label{fig:speedup} Convergence at Different Mobility Speed.}
\end{figure}

Vehicle mobility directly determines the frequency and efficiency of cache-based model spreading and aggregation. 
So we evaluate the performance of \DFC at different vehicle speeds. We fix cache size at 10, $\tau_{max}=10$ with  non-i.i.d. data distribution. We take the our previous speed $v=v_0=13.89m/s$ and $K=30$ local updates as the base, named as speedup x1. To speedup, $v$ increases while $K$ reduces to keep the fair comparison under the same wall clock. For instance, for speedup x3, $v=3v_0$ and $K=10$. Results in Fig. \ref{fig:speedup} show that when the mobility speed increases, although the number of local updates decreases, the spread of all models in the whole vehicle network is boosted, thus disseminating local models more quickly among all vehicles leading to faster model convergence. 

\subsection{Experiments with Grouped Mobility Patterns and Data Distributions}
In practice, vehicle mobility patterns and local data distributions may naturally form groups. For example, a vehicle may mostly move within its home area, and collect data specific to that area. Vehicles within the same area meet with each others frequently, but have very similar data distributions. A fresh model of a vehicle in the same area is not as valuable as a slightly older model of a vehicle from another area. So model caching should not just consider model freshness, but should also take into account the coverage of group-based data distributions. For those scenarios, we develop a Group-based (GB) caching algorithm as Algorithm~\ref{alg: GB}. More specifically, knowing that there are $m$ distribution groups, one should maintain balanced presences of models from different groups. One straightforward extension of the LRU caching algorithm is to partition the cache into $m$ sub-caches, one for each group. Each sub-cache is updated by local models from its associated group, using the LRU policy. 
  
\begin{algorithm}
	\caption{Group Based  Cache Update (GB Update)} \label{alg: GB}
 \textbf{Input:} Current cache $\mathcal{C}_i(t)$, Cache of device $j$: $\mathcal{C}_j(t)$, Model $x_j(t)$ from device $j$, Current time $t$, Cache size $\mathcal{C}_{max}$, Tolerance of staleness $\tau_{max}$, Device group mapping list $\mathcal{A} = \{\mathcal{A}_1,\mathcal{A}_2,...,\mathcal{A}_n\}$, here $\mathcal{A}_i\in \{1,2,...,m\},\forall i\in [N]$. Group cache size list $\mathcal{R} = \{r_1,r_2,...,r_m\}$, here $r_{\mathcal{A}_i}$ is number of cache slots reserved for models from devices belong to group $\mathcal{A}_i$.
 \\ Here the model $x(i,t')$ is local updated at time $t'$, based on car $i$'s dataset.
\begin{algorithmic}[1]
\Statex\Function{Group\_prune\_cache}{$\mathcal{C}_i(t),\mathcal{R},\mathcal{A}$}:
\State Create empty lists of group $\mathcal{L}_1,\mathcal{L}_2,...,\mathcal{L}_m,$ 
        \For {$x_k(\tau)\in \mathcal{C}_i(t)$}\label{line: group cached model}
        \State Put $x_k(\tau)$ into $\mathcal{L}_{\mathcal{A}_k}$
        \EndFor
        \State Empty $\mathcal{C}_i(t)$
        \For{$k=1,2,...,m$}\label{line:group sort}
        \State Sort models in $\mathcal{L}_k$ by $\tau$ descending.
        \State Take and put first $r_k$ models in $\mathcal{L}_k$ into $\mathcal{C}_i(t)$
        \EndFor
        \State \Return $\mathcal{C}_i(t+1)$
\EndFunction
\newline
\setcounter{ALG@line}{0}
\Statex\textbf{Main Process:}

\For {$x_k(\tau)\in \mathcal{C}_i(t)$ or $\mathcal{C}_j(t)$}
\If {$t-\tau\geq\tau_{max}$}
\State Remove $x_k(\tau)$ from related cache $\mathcal{C}_i(t)$ or $\mathcal{C}_j(t)$
\EndIf
\EndFor
\State Add or replace $x_j(t)$ into $\mathcal{C}_i(t)$
\For{$x_k(\tau)\in \mathcal{C}_j(t)$}
\If{$x_k\notin \mathcal{C}_i(t)$}
\State Add $x_k(\tau)$ into $\mathcal{C}_i(t)$
\Else
\State Retrieve $x_k(\tau')\in \mathcal{C}_i(t)$
\If{$\tau>\tau'$}
\State Replace $x_k(\tau)$ with $x_k(\tau')$ into $\mathcal{C}_i(t)$
\EndIf
\EndIf
\EndFor
		
	\State	$\mathcal{C}_i(t+1)\gets$ \Call{GROUP\_PRUNE\_CACHE}{$\mathcal{C}_i(t),\mathcal{R},\mathcal{A}$}\label{line:group_prune_cache}
	\end{algorithmic}
\textbf{ Output:} $\mathcal{C}_i(t+1)$
\end{algorithm}

\begin{figure}
\centering
\includegraphics[width=1\textwidth]{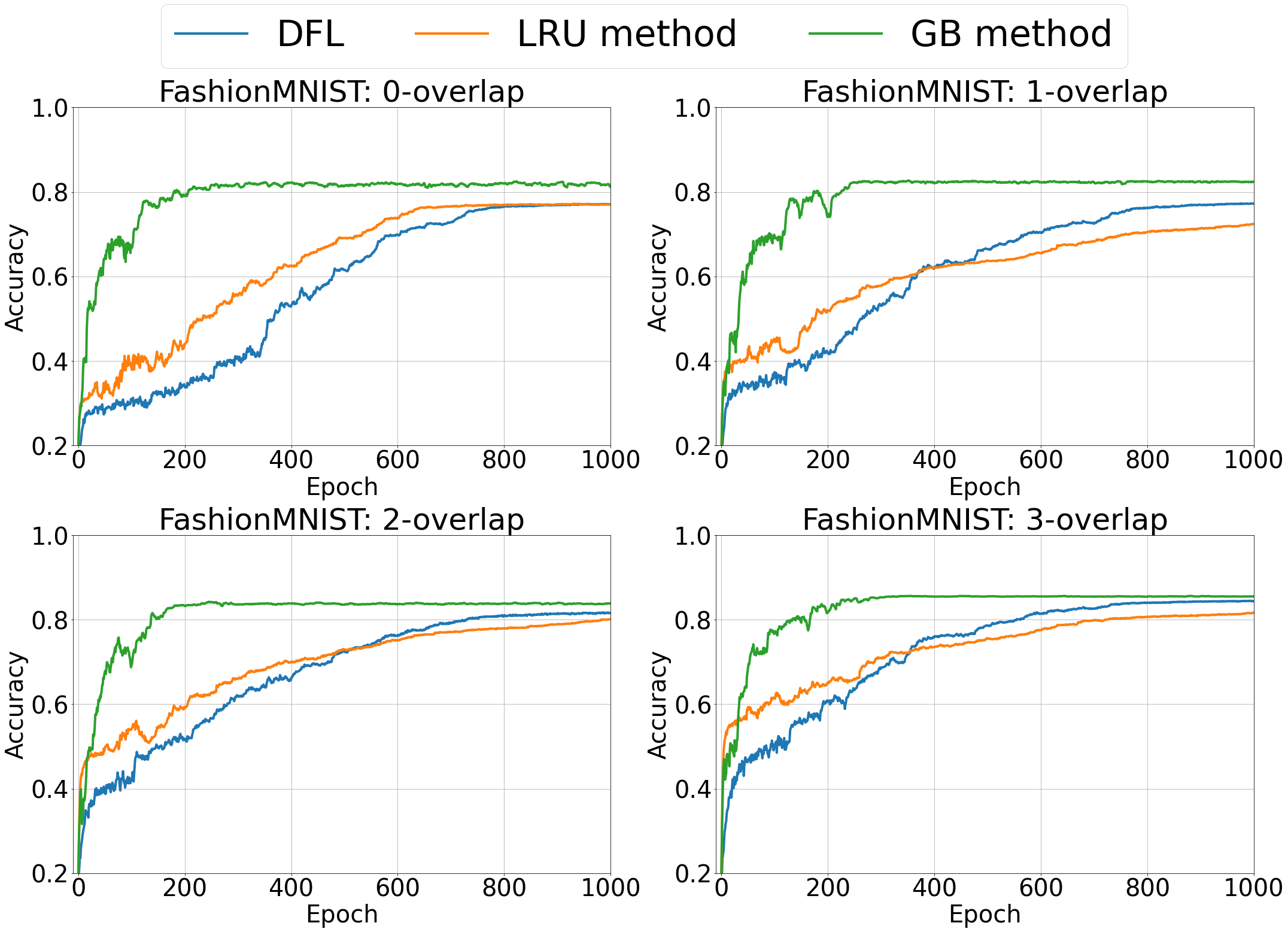}
\caption{\label{fig:Fashionmnist_GB_update} Group-based LRU Cache Update Performance under Different Data Distribution Overlaps.}
\end{figure}

We now conduct a case study for group-based cache update. 
As shown in Fig. \ref{fig:3_area}, the whole Manhattan road network is divided into 3 areas, downtown, mid-town and uptown. Each area has 30 area-restricted vehicles that randomly moves within that area, and 3 or 4 free vehicles that can move into any area. 
We set 4 different area-related data distributions: Non-overlap, 1-overlap, 2-overlap, 3-overlap. $n$-overlap means the number of shared label classes between areas is $n$. 
We use the same non-i.i.d shards method in the previous section to allocate data points to the vehicles in the same area. 
On each vehicle, we evenly divide the cache for the three areas. We evaluate our proposed GB cache method on FashionMNIST. As shown in Fig. \ref{fig:Fashionmnist_GB_update}, while vanilla LRU converges faster at the very beginning, it cannot  outperform DFL at last. However, the GB cache update method can solve the problem of LRU update and outperform DFL under different overlap settings. 

\subsection{Discussions}
Decentralized FL (DFL) has been increasingly applied in vehicular networks, leveraging existing frameworks like vehicle-to-vehicle (V2V) communication \cite{yuan2024decentralized}. V2V FL facilitates knowledge sharing among vehicles and has been explored in various studies \cite{samarakoon2019distributed,pokhrel2020decentralized,yu2020proactive,chen2021bdfl, barbieri2022decentralized, su2022boost}. \citet{samarakoon2019distributed} studied optimized joint power and resource allocation for ultra-reliable low-latency communication (URLLC) using FL. \citet{su2022boost} introduced DFL with Diversified Data Sources to address data diversity issues in DFL, improving model accuracy and convergence speed in vehicular networks. None of the previous studies explored model caching on vehicles.
Convergence of asynchronous federated optimization was studied in \citet{DBLP:journals/corr/abs-1903-03934}. 
Their analysis focused on pairwise model aggregation between an agent and the parameter server, does not cover decentralized model aggregation with stale cached models in our proposed framework.  
\\

\section{Related Work}\label{sec:related}
Decentralized FL (DFL) has been increasingly applied in vehicular networks, leveraging existing frameworks like vehicle-to-vehicle (V2V) communication \cite{yuan2024decentralized}. V2V FL facilitates knowledge sharing among vehicles and has been explored in various studies \cite{samarakoon2019distributed,pokhrel2020decentralized,yu2020proactive,chen2021bdfl, barbieri2022decentralized, su2022boost}. \citet{samarakoon2019distributed} studied optimized joint power and resource allocation for ultra-reliable low-latency communication (URLLC) using FL. 

\citet{su2022boost} introduced DFL with Diversified Data Sources to address data diversity issues in DFL, improving model accuracy and convergence speed in vehicular networks.
None of the previous studies explored model caching on vehicles.
Convergence of asynchronous federated optimization was studied in \citet{DBLP:journals/corr/abs-1903-03934}. 
Their analysis focused on pairwise model aggregation between an agent and the parameter server, does not cover decentralized model aggregation with stale cached models in our proposed framework.

\section{Conclusion \& Future Work}\label{Conclusion}
In this paper, we developed \DFC, a novel decentralized Federated Learning framework that leverages on model caching on mobile agents for fast and even model spreading. We theoretically analyzed the convergence of \DFC. Through extensive case studies in a vehicle network, we demonstrated that \DFC  significantly outperforms DFL without model caching, especially for agents with non-i.i.d data distributions. We employed only simple model caching and aggregation algorithms in the current study. We will investigate more refined model caching and aggregation algorithms customized for different agent mobility patterns and non-i.i.d. data distributions.

\clearpage

\bibliography{refs,refs_app}
\bibliographystyle{plainnat}

\onecolumn

\appendix
\section{Convergence Analysis}
We now theoretically investigate the impact of caching, especially the staleness of cached models, on DFL model convergence. We introduce some definitions and assumptions. 
\begin{definition}[Smoothness]
A differentiable function $f$  is $L$-smooth if for $\forall x,y,$ $f(y)-f(x)\leq \langle  \nabla f(x),y-x \rangle+\frac{L}{2}||y-x||^2$, where $L>0$.
\end{definition}
\begin{definition}[Bounded Variance]
  There exists a constant  $\varsigma>0$ such that the global variability of the local gradient of the loss function is bounded $||\nabla F_j(x)-\nabla F(x)||^2\leq \varsigma^2,\forall j \in[N],x\in \mathbb{R}^d$.  
\end{definition}
\begin{definition}[L-Lipschitz Continuous Gradient]
  There exists a constant $L>0$, such that $||\nabla F(x)-\nabla F(x')||\leq L'||x-x'||,\forall x,x'\in \mathbb{R}^d.$  
\end{definition}

\begin{theorem}
Assume that $F$ is $L$-smooth and convex, and each agent executes $K$ local updates before meeting and exchanging models, after that, then does model aggregation. We also assume bounded staleness $\tau < \tau_{max}$, as the kick-out threshold. Furthermore, we assume, $\forall x\in \mathbb{R}^d, i \in [N]$, and $\forall z\sim \mathcal{D}^i, ||\nabla f(x;z)||^2\leq V,||\nabla g_{x'}(x;z)||^2\leq V,\forall x'\in \mathbb{R}^d$, and $\nabla F$ satisfies L-Lipschitz Continuous Gradient. For any small constant $\epsilon>0$, if we take $\rho>0$, and satisfying $-(1+2\rho+\epsilon)V+(\rho^2-\frac{\rho}{2})||x(t,k-1)-x(t)||^2\geq 0,\forall x(t,k-1),x(t)$, after $T$ global epochs, Algorithm 1 converges to a critical point: 
\begin{align}
\begin{split}
    \min_{t=0}^{T-1} \mathbb{E}||\nabla F(x(t))||^2\leq&\frac{\tau_{max}}{\epsilon\eta C_1KT} \mathbb{E}[F(x(0))-F(x_{M(T)}(T))]\nonumber
    +\mathcal{O}(\frac{\eta \tau_{max} K^2}{\epsilon C_1T})\\
    \leq &\mathcal{O}(\frac{\tau_{max}}{\epsilon\eta C_1KT}) +\mathcal{O}(\frac{\eta\rho  K^2}{\epsilon C_1}). 
\end{split}
\end{align}
\end{theorem}
\subsection{Proof}
Similar to Theorem 1 in~\cite{DBLP:journals/corr/abs-1903-03934}, we can get the boundary before and after $K$ times local updates on $j$-th device, $\forall j\in [N]$:
\begin{align}\nonumber
    \mathbb{E}[F(\Tilde{x}_j(t))-F(x_j(t))] &=\mathbb{E}[F(x_j(t,K))-F(x_j(t,0))]\\
    &\leq -\eta \epsilon \sum_{k=0}^{K-1}\mathbb{E}||\nabla F(x_j(t,k))||^2+\eta^2\mathcal{O}(\rho K^3V).
\end{align}
For any epoch $t$, 
we find the index $M(t)$ of the agent whose model is the ``worst", i.e., $M(t) = \arg\max_{j \in [N]}\{F(x_j(t))\}$,
then we can get,
\begin{align}\nonumber
        \mathbb{E}[F&(x_i(t+1))-F(x_*)]\leq \mathbb{E}[F(x_{M(t+1)}(t+1))-F(x^*)]\\\nonumber
        \leq& \mathbb{E}\Big[F\Big(\frac{1}{|C_{M(t+1)}(t)|}\sum_{j,\tau\in C_{M(t+1)}(t)}\Tilde{x}_j(\tau)\Big)\Big]-F(x^*)\\\nonumber
        \leq& \mathbb{E}\Big[\frac{1}{|C_{M(t+1)}(t)|}\sum_{j,\tau\in C_{M(t+1)}(t)}F(\Tilde{x}_j(\tau))\Big]-F(x^*)\\\nonumber
        \leq& \mathbb{E}\Big[\frac{1}{|C_{M(t+1)}(t)|}\sum_{j,\tau\in C_{M(t+1)}(t)}F(x_j(\tau))\Big]-F(x^*)+\eta^2\mathcal{O}(\rho K^3V)\\\nonumber
        &-\frac{\epsilon\eta}{|C_{M(t+1)}(t)|}\sum_{j,\tau\in C_{M(t+1)}(t)} \sum_{k=0}^{K-1}\mathbb{E}||\nabla F(x_j(\tau,k))||^2\\\nonumber
        \leq& \mathbb{E}\Big[\frac{1}{|C_{M(t+1)}(t)|}\sum_{\tau = t-\tau_{max}+1}^{t}\mathrlap{\Big(}\sum_{\substack{x_j(\tau)\in C_{M(t)}(t)\\x_j(\tau)\neq x_{M(\tau)}(\tau)}}\mathllap{F(}x_j(\tau))+\sum_{x_{M(\tau)}(\tau)\in C_{M(t)}(t)}\mathllap{F(}x_{M(\tau)}(\tau))\Big)\Big]\\
        &- \nabla F -F(x^*)+\eta^2\mathcal{O}(\rho K^3V).
\end{align}
here $\nabla F=\frac{\epsilon\eta}{|C_{M(t+1)}(t)|}\sum_{j,\tau\in C_{M(t+1)}(t)} \sum_{k=0}^{K-1}\mathbb{E}||\nabla F(x_j(\tau,k))||^2 $.\\
We can easy know, for given $\tau,\forall j\neq M(\tau)$, we have $F(x_{M(\tau)}(\tau))> F(x_j(\tau))  $, then, 
\begin{equation}
    \mathbb{E}\Big[\sum_{\substack{x_j(\tau)\in C_{M(t)}(t)\\x_j(\tau)\neq x_{M(\tau)}(\tau)}}F(x_j(\tau))\Big]=\Big(\sum_{\substack{x_j(\tau)\in C_{M(t)}(t)\\x_j(\tau)\neq x_{M(\tau)}(\tau)}}\cdot 1\Big) h(\tau)F(x_{M(\tau)}(\tau)) . 
\end{equation}
Here $0<h(\tau)<1 $,
so we rearrange the equation above, we can get:
\begin{align}\nonumber
    \mathbb{E}[F&(x_{M(t+1)}(t+1))-F(x^*)]\\\nonumber
        \leq& \mathbb{E}\Big[\sum_{\tau = t}^{t-\tau_{max}+1}\frac{\sum_{\substack{x_j(\tau)\in C_{M(t)}(t)\\x_j(\tau)\neq x_{M(\tau)}(\tau)}}\cdot h(\tau)+\sum_{x_{M(\tau)}(\tau)\in C_{M(t)}(t)}\cdot 1}{|C_{M(t+1)}(t)|}F(x_{M(\tau)}(\tau))\Big]\\\nonumber
        &- \nabla F -F(x^*)+\eta^2\mathcal{O}(\rho K^3V)\\\nonumber
        =& \mathbb{E}\Big[\sum_{\tau = t}^{t-\tau_{max}+1}\frac{\sum_{\substack{x_j(\tau)\in C_{M(t)}(t)\\x_j(\tau)\neq x_{M(\tau)}(\tau)}}\cdot h(\tau)+\sum_{x_{M(\tau)}(\tau)\in C_{M(t)}(t)}\cdot 1}{|C_{M(t+1)}(t)|}F(x_{M(\tau)}(\tau))\Big]\\\nonumber
        &- \nabla F-F(x^*)+\eta^2\mathcal{O}(\rho K^3V)\\\nonumber
        =& \mathbb{E}\Big[\sum_{\tau = t}^{t-\tau_{max}+1}\frac{\sum_{\substack{x_j(\tau)\in C_{M(t)}(t)\\x_j(\tau)\neq x_{M(\tau)}(\tau)}}\cdot h(\tau)+\sum_{x_{M(\tau)}(\tau)\in C_{M(t)}(t)}\cdot 1}{\sum_{\tau = t}^{t-\tau_{max}+1}(\sum_{\substack{x_j(\tau)\in C_{M(t)}(t)\\x_j(\tau)\neq x_{M(\tau)}(\tau)}}\cdot 1+\sum_{x_{M(\tau)}(\tau)\in C_{M(t)}(t)}\cdot 1)}F(x_{M(\tau)}(\tau))\Big]\\
        &- \nabla F -F(x^*)+\eta^2\mathcal{O}(\rho K^3V).
\end{align}
We define the ``worst" model on all agents over the time period of $[t-\tau_{max}+1, t]$ as 
\[ \mathcal{T}(t,\tau_{max})= \arg\max_{t \in [t-\tau_{max}+1, t]} \{F(x_{M(t)}(t))\}.\]
Then we can get,
\begin{align}\nonumber
    \mathbb{E}[F&(x_{M(t+1)}(t+1))-F(x^*)]\\\nonumber
        \leq& \mathbb{E}\Big[\sum_{\tau = t}^{t-\tau_{max}+1}\frac{\sum_{\substack{x_j(\tau)\in C_{M(t)}(t)\\x_j(\tau)\neq x_{M(\tau)}(\tau)}}\cdot h(\tau)+\sum_{x_{M(\tau)}(\tau)\in C_{M(t)}(t)}\cdot 1}{\sum_{\tau = t}^{t-\tau_{max}+1}(\sum_{\substack{x_j(\tau)\in C_{M(t)}(t)\\x_j(\tau)\neq x_{M(\tau)}(\tau)}}\cdot 1+\sum_{x_{M(\tau)}(\tau)\in C_{M(t)}(t)}\cdot 1)}\\\nonumber
        &\cdot F(x_{(\mathcal{T}(t,\tau_{max}))}(\mathcal{T}(t,\tau_{max})))\Big]-\nabla F -F(x^*)+\eta^2\mathcal{O}(\rho K^3V)\\\nonumber
        =& \mathbb{E}\Big[\Big(1-\sum_{\tau = t}^{t-\tau_{max}+1}\frac{\sum_{\substack{x_j(\tau)\in C_{M(t)}(t)\\x_j(\tau)\neq x_{M(\tau)}(\tau)}}\cdot (1-h(\tau))}{|C_{M(t+1)}(t)|}\Big)F(x_{(\mathcal{T}(t,\tau_{max}))}(\mathcal{T}(t,\tau_{max})))\Big]\\\nonumber
        &- \nabla F -F(x^*)+\eta^2\mathcal{O}(\rho K^3V)\\\nonumber
        =& \mathbb{E}\Big[\Big(1-\xi(t,\tau_{max})\Big)F(x_{(\mathcal{T}(t,\tau_{max}))}(\mathcal{T}(t,\tau_{max})))\Big]- \nabla F -F(x^*)+\eta^2\mathcal{O}(\rho K^3V)\\
        \leq& \mathbb{E}\Big[F(x_{(\mathcal{T}(t,\tau_{max}))}(\mathcal{T}(t,\tau_{max})))\Big]- \nabla F -F(x^*)+\eta^2\mathcal{O}(\rho K^3V).
\end{align}
Here,
$\xi(t,\tau_{max}) = \sum_{\tau = t}^{t-\tau_{max}+1}\frac{\sum_{\substack{x_j(\tau)\in C_{M(t)}(t)\\x_j(\tau)\neq x_{M(\tau)}(\tau)}}\cdot (1-h(\tau))}{|C_{M(t+1)}(t)|}\in [0,1)$, so we have,
\begin{equation}
    \mathbb{E}[F(x_{M(t+1)}(t+1))-F(x_{(\mathcal{T}(t,\tau_{max}))}(\mathcal{T}(t,\tau_{max})))]
        \leq -\nabla F +\eta^2\mathcal{O}(\rho K^3V).
\end{equation}
Then,
\begin{align}\label{eq:bound1_0}\nonumber
    \mathbb{E}[F&(x_{M(t+1)}(t+1))-F(x_{(\mathcal{T}(t,\tau_{max}))}(\mathcal{T}(t,\tau_{max})))]\\\nonumber
        &\leq -\nabla F +\eta^2\mathcal{O}(\rho K^3V) \\
        &= -\frac{\epsilon\eta}{|C_{M(t+1)}(t)|}\sum_{j,\tau\in C_{M(t+1)}(t)}\sum_{k=0}^{K-1} ||\nabla F(x_j(\tau,k))||^2+\eta^2\mathcal{O}(\rho K^3V).  
\end{align}
By taking use of the property of L-Lipschitz Continuous Gradient, we can get,
\begin{align}\nonumber
    ||\nabla F(x_j(\tau,k))-\nabla F(x_j(\tau,k-1))||&\leq L'||x_j(\tau,k)-x_j(\tau,k-1)||\\\nonumber
    &\leq \eta L'||\nabla g_{x_j(\tau)}(x_j(\tau,k-1))||\\
    &\leq\eta L' \sqrt{V}.
\end{align}
Then, we can get, $\forall k\in [1,K]$, $||\nabla F(x_j(\tau,k))||^2\geq||\nabla F(x_j(\tau,k-1))||^2+\eta^2 L'^2V$, then
\begin{equation}
    ||\nabla F(x_j(\tau,k))||^2\geq||\nabla F(x_j(\tau))||^2+\eta^2 L'^2 kV.
\end{equation}
So we have,
\begin{align}\nonumber
    \sum_{j,\tau\in C_{M(t+1)}(t)}\sum_{k=0}^{K-1} &||\nabla F(x_j(\tau,k))||^2\\ \nonumber
    &\geq \sum_{j,\tau\in C_{M(t+1)}(t)}K||\nabla F(x_j(\tau))||^2+\eta^2 L'^2 K(K-1)V\\
    &\geq \sum_{j,\tau\in C_{M(t+1)}(t)}K||\nabla F(x_j(\tau))||^2+\eta^2\mathcal{O}( K^2V).
\end{align}
We assume the ratio of gradient expectation on the global distribution to gradient expectation on local data distribution of any device is bounded, i.e., 
\begin{align}
    \mathbb{E}[||\nabla F(x_j(t))||^2]\geq C_1\mathbb{E}[||\nabla F(x(t))||^2],
\end{align}
So we get,
\begin{align}\nonumber
    \sum_{j,\tau\in C_{M(t+1)}(t)}&\sum_{k=0}^{K-1} \mathbb{E}[||\nabla F(x_j(\tau,k))||^2] \\\nonumber
    \ge&C_1 K  |C_{M(t+1)}(t)|\min_{\tau=t-\tau_{max}+1}^{t} \mathbb{E}[||\nabla F(x(\tau))||^2]\\
    &+\eta^2|C_{M(t+1)}(t)|\mathcal{O}(K^2V),
\end{align}
Inequality (\ref{eq:bound1_0}) becomes:  
\begin{align}\label{eq:bound2}\nonumber
    \mathbb{E}&[F(x_{M(t+1)}(t+1))-F(x_{(\mathcal{T}(t,\tau_{max}))}(\mathcal{T}(t,\tau_{max})))]\\
                &\leq -\epsilon\eta C_1K \min_{\tau=t-\tau_{max}+1}^{t} ||\nabla F(x(\tau))||^2+\eta^2\mathcal{O}(\rho K^3V).
\end{align}  
By rearranging the terms, we have,
\begin{align}\nonumber\label{eq:bound1_appendix}
        \min_{\tau=t-\tau_{max}+1}^{t} ||\nabla F(x(\tau))||^2
        \leq& (\frac{1}{\epsilon\eta C_1K })\mathbb{E}[F(x_{(\mathcal{T}(t,\tau_{max}))}(\mathcal{T}(t,\tau_{max})))\\
        &-F(x_{M(t+1)}(t+1))]+\mathcal{O}(\frac{\eta\rho K^2V}{C_1\epsilon}).
\end{align}
We iteratively construct a time sequence  $\{T'_0,T'_1,T'_2,...,T'_{N_T}\}\subseteq \{0,1,...,T-1\}$ in the backward fashion so that 
 \begin{eqnarray*}
    &T'_{N_T} &= T-1;\\
    &T'_i &= \mathcal{T}(T'_{i+1},\tau_{max})-1, \quad 1 \le i \le N_T-1;\\    
    &T'_0 &= 0.
\end{eqnarray*}
From the equation above, we can also get the inequality about $N_T$, 
\begin{equation}
    \frac{T}{\tau_{max}}\leq N_T\leq T.
\end{equation}
Applying inequality~(\ref{eq:bound1_appendix}) at all time instances $\{T'_0,T'_1,T'_2,...,T'_{N_T}\}$, after T global epochs, we have,
\begin{align}\label{eq:gradientbound_app}\nonumber
    \min_{t=0}^{T-1} \mathbb{E}[&||\nabla F(x(t))||^2]\leq \frac{1}{N_T} \sum_{t=T'_0}^{T'_N}\min_{\tau=t-\tau_{max}+1}^{t} ||\nabla F(x(\tau))||^2\\\nonumber
        \leq& \frac{1}{\epsilon\eta C_1K N_T}\sum_{t=T'_0}^{T'_N} \mathbb{E}[F(x_{M_{T(t,\tau_{max})}}(T(t,\tau_{max})))-F(x_{M(t+1)}(t+1))]+\mathcal{O}(\frac{\eta\rho K^2V}{\epsilon C_1})\\\nonumber
        \leq& \frac{1}{\epsilon\eta C_1KN_T} \mathbb{E}[F(x_{M(T'_0)}(T'_0))-F(x_{M(T'_{N_T}+1)}(T'_{N_T}+1))]+\mathcal{O}(\frac{\eta\rho K^2V}{\epsilon C_1})  \\\nonumber
        \leq& \frac{1}{\epsilon\eta C_1KN_T} \mathbb{E}[F(x_{M(0)}(0))-F(x_{M(T)}(T))]+\mathcal{O}(\frac{\eta\rho K^2V}{\epsilon C_1} )\\\nonumber
        \leq& \frac{1}{\epsilon\eta C_1KN_T} \mathbb{E}[F(x(0))-F(x_{M(T)}(T))]+\mathcal{O}(\frac{\eta\rho K^2V}{\epsilon C_1})\\\nonumber
        \leq& \frac{\tau_{max}}{\epsilon\eta C_1KT} \mathbb{E}[F(x(0))-F(x_{M(T)}(T))]+\mathcal{O}(\frac{\eta\rho  K^2V}{\epsilon C_1}) \\\nonumber
        \leq& \frac{\tau_{max}}{\epsilon\eta C_1KT} \mathbb{E}[F(x(0))-F(x(T)]+\mathcal{O}(\frac{\eta\rho K^2V}{\epsilon C_1})  \\
         \leq& \mathcal{O}(\frac{\tau_{max}}{\epsilon\eta C_1KT}) +\mathcal{O}(\frac{\eta\rho  K^2}{\epsilon C_1}) .
\end{align}
With the results in~(\ref{eq:gradientbound_app}) and by leveraging Theorem 1 in \citet{yang2021achieving}, Algorithm~1 converges to a critical point after $T$ global epochs.
\section{Experimental details}
\subsection{Data distribution in Group-based LRU Cache Experiment}
\subsubsection{\texorpdfstring{$n$}{n}-overlap}
For MNIST, FashionMNIST with 10 classes labels, we allocate labels into 3 groups as:\\
\textbf{Non-overlap:}\\ Area 1: (0,1,2,3), Area 2: (4,5,6), Area 3: (7,8,9)\\
\textbf{1-overlap:}\\ Area 1: (9,0,1,2,3), Area 2: (3,4,5,6), Area 3: (6,7,8,9)\\
\textbf{2-overlap:}\\ Area 1: (8,9,0,1,2,3), Area 2: (2,3,4,5,6), Area 3: (5,6,7,8,9)\\
\textbf{3-overlap:}\\ Area 1: (7,8,9,0,1,2,3), Area 2: (1,2,3,4,5,6), Area 3: (4,5,6,7,8,9)\\
\subsection{Computing infrastructure}
For MNIST and FashionMNIST experiments, we use 10 computing nodes, each with 10 CPUs, model: Lenovo SR670, to simulate DFL on 100 vehicles.
CIFAR-10 results are obtained from 1 compute node with 5 CPUs and 1 A100 NVIDIA GPU, model: SD650-N V2. 
They are all using Ubuntu-20.04.1 system with pre-installed openmpi/intel/4.0.5 module, and 100 Gb/s network speed.
\subsection{Results with variation}
As our metric is the average test accuracy over 100 devices, we also provide the variation of the average test accuracy in our experimental results as following,
\begin{figure}[H]
\centering
\includegraphics[width=1.0\textwidth]{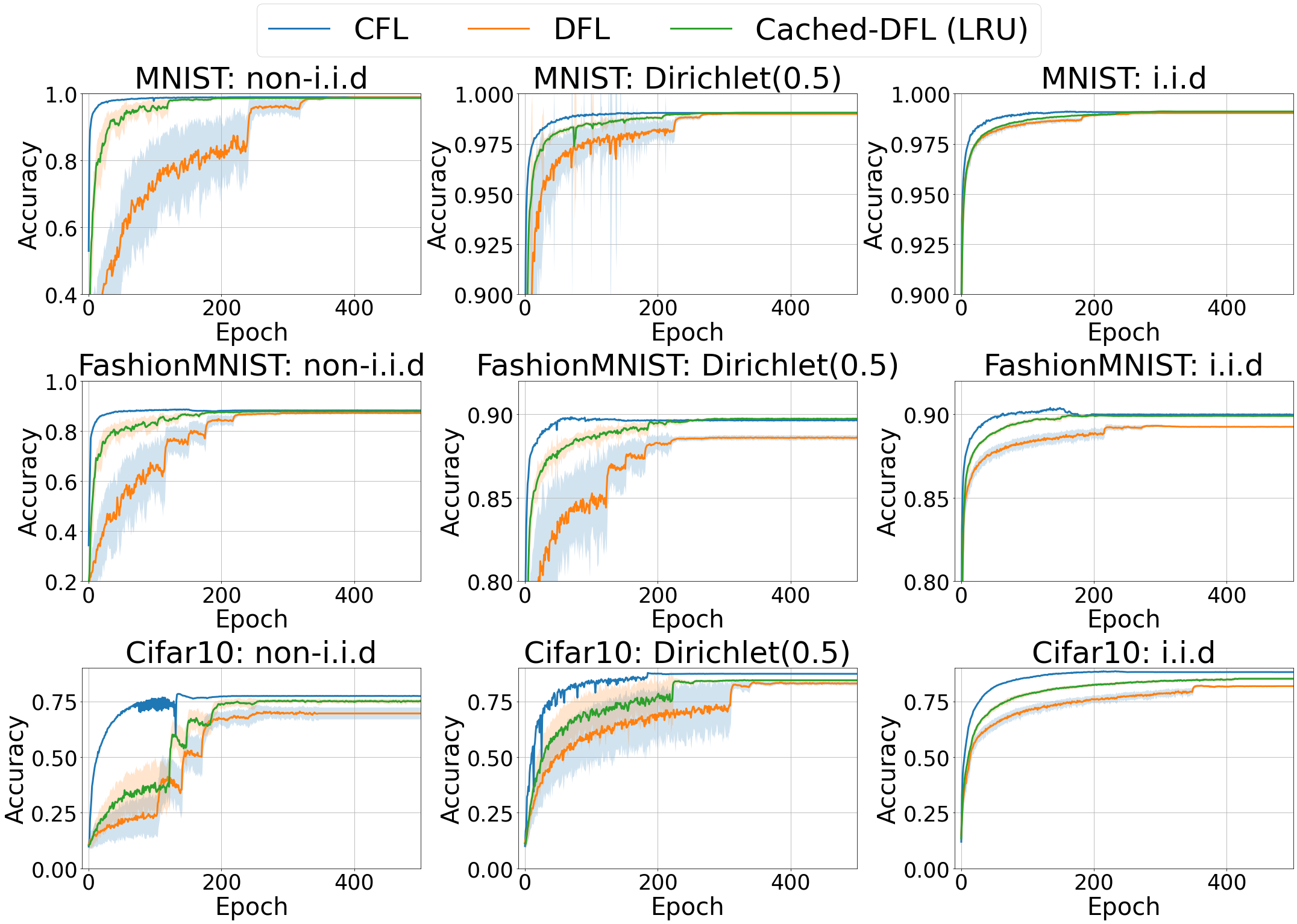}
\caption{\label{fig:cache_non_cache_var} DFL with Caching vs. DFL without Caching with Variation.}
\end{figure}

\begin{figure}[H]
\centering
\includegraphics[width=1.0\textwidth]{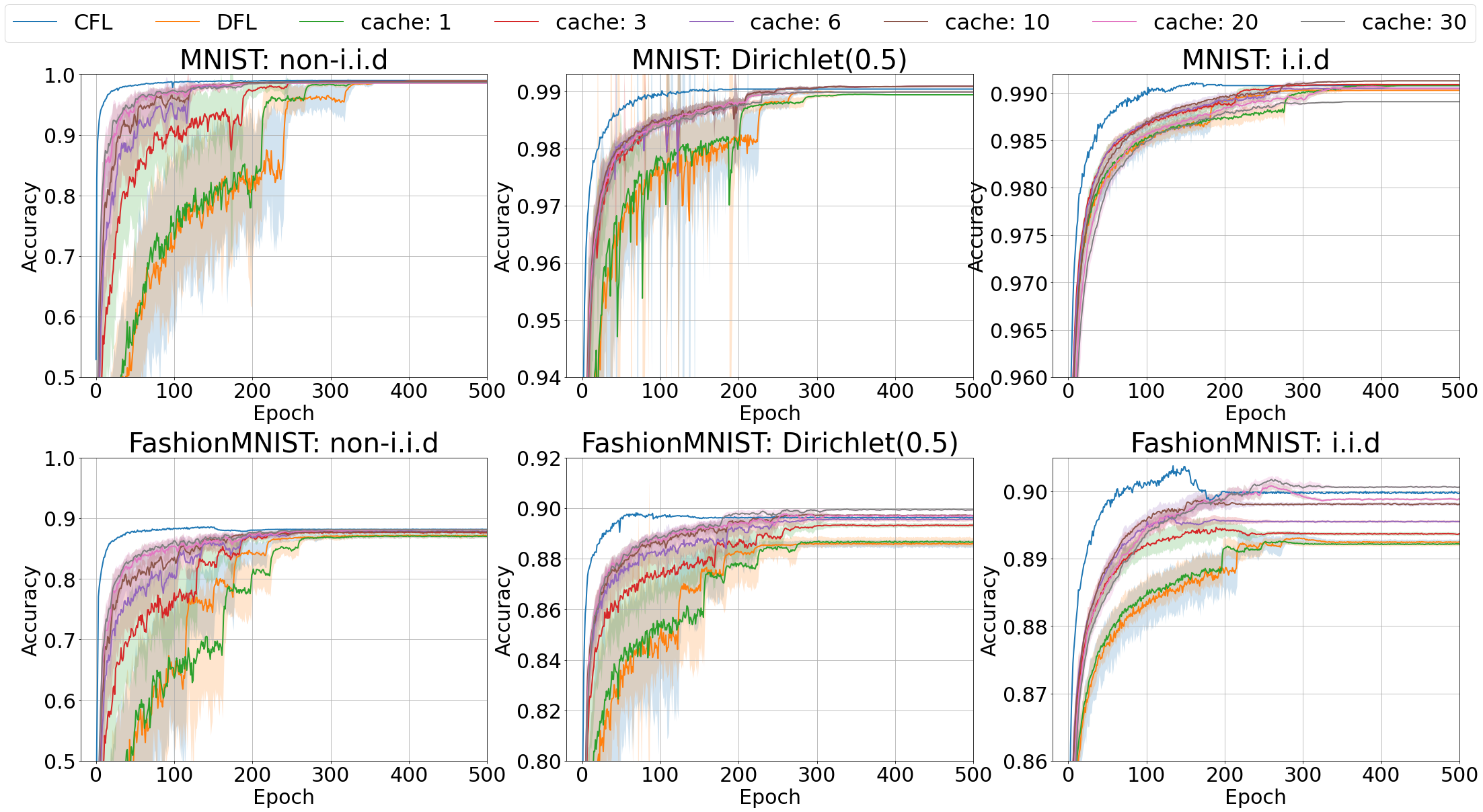}
\caption{\label{fig:different_cache_var} DFL with LRU at Different Cache Sizes with Variation.}
\end{figure}
\begin{figure}[H]
\centering
\includegraphics[width=1.0\textwidth]{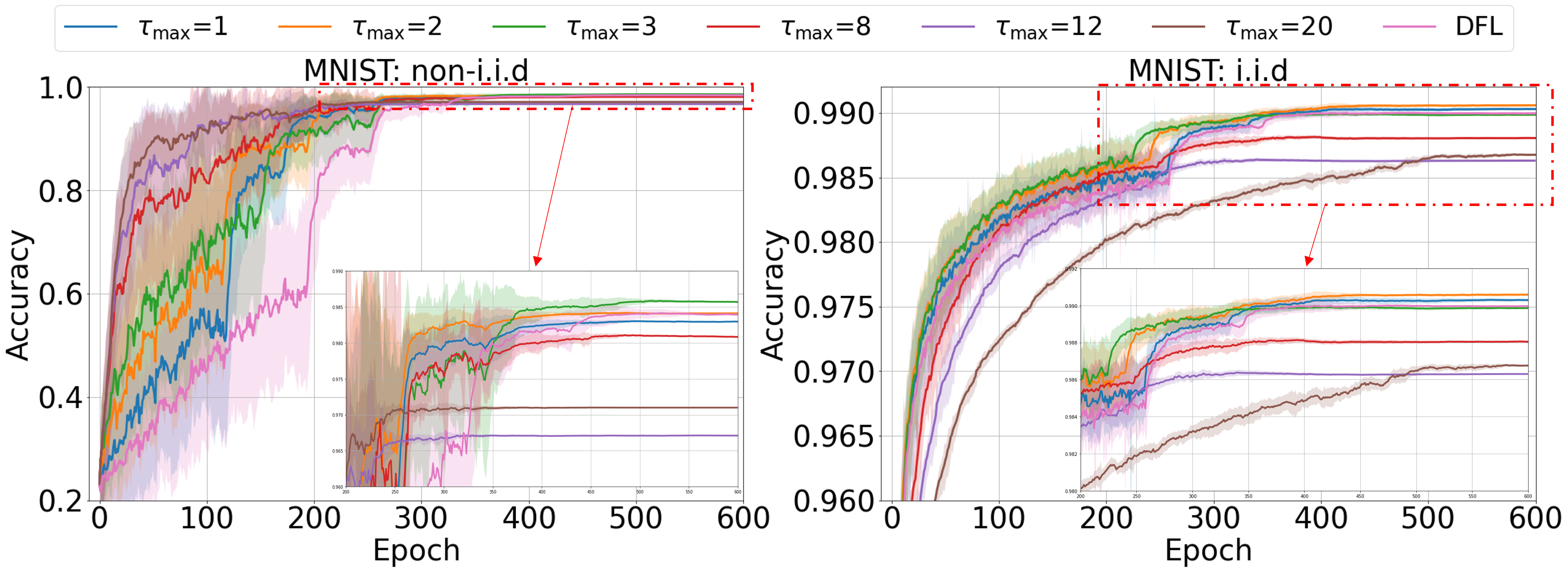}
\caption{\label{fig:different_kick_out_var} Impact of  $\tau_{max}$ on Model Convergence with Variation.}
\end{figure}

\begin{table}
\centering
  \caption{Average number and average age of cached models with different $\tau_{max}$ and different epoch time: 30s, 60s, 120s. Columns for different $\tau_{max}$, and rows for different epoch time. There are four sub-rows for each epoch time, with the first two sub-rows be the average number of cached models and variation, the second two sub-rows be their average age and variation.}
  \label{tab:cache_size_age_tradeoff_var}
  \begin{tabular}{|c|r|r|r|r|r|r|r|r|}
    \hline
     ~&$\tau_{max}$& 1&2&3&4&5&10&20 \\
\hline
    \multirow{4}*{ 30s} &num &0.8549&1.6841&2.6805&3.8584&5.2054&14.8520&44.8272\\
    \cline{2-9}
    ~&$var_n$&0.0321&0.0903&0.1941&0.3902&0.6986&6.2270&42.8144\\\cline{2-9}
    ~&age&0.0000&0.4904&1.0489&1.6385&2.2461&5.4381&11.5735\\ \cline{2-9}
    ~&$var_a$&0.0000&0.0052&0.0116&0.0203&0.0320&0.1577&1.0520\\ \hline
    \multirow{4}*{ 60s} &num&1.6562&3.8227&6.4792&10.3889&14.0749&43.0518&90.2576\\
    \cline{2-9}
    ~&$var_n$&0.0865&0.3594&0.9925&2.824&5.1801&32.8912&55.5829\\\cline{2-9}
    ~&age&0.000&0.5593&1.1604&1.8075&2.4396&5.5832&9.7954\\\cline{2-9}
    ~&$var_a$&0.0000&0.0035&0.0089&0.0188&0.0251&0.1448&0.8375\\ \hline
    \multirow{4}*{ 120s}  &num&3.6936&9.9149&18.5016&31.3576&35.1958&90.2469&98.5412\\
    \cline{2-9}
    ~&$var_n$&0.3166&2.3825&5.9696&21.7408&56.8980&28.8958&29.9676\\\cline{2-9}
    ~&age&0.0000&0.6189&1.2715&1.9194&1.5081&4.7131&5.2357\\ \cline{2-9}
    ~&$var_a$&0.0000&0.0028&0.0058&0.0212&0.9742&0.1249&0.1808\\ \hline
\end{tabular}
\end{table}

\begin{figure}[H]
\centering
\includegraphics[width=1.0\textwidth]{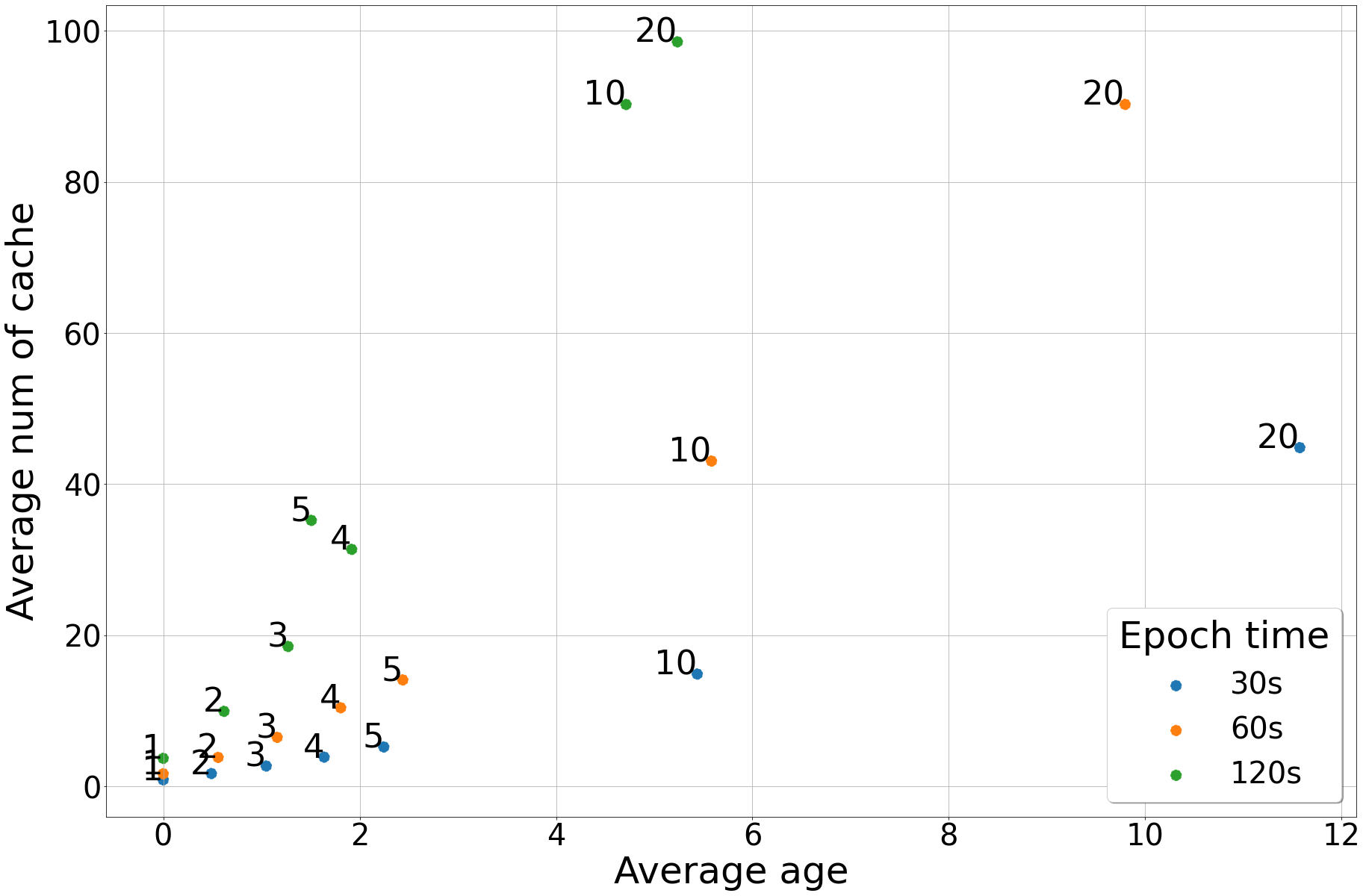}
\caption{\label{fig:cache_size_age_tradeoff_var} Relation Between Average Number and Average Age of Cached Models with Different $\tau_{max}$ (number close to the dot), with Unlimited Cache Size.}
\end{figure}
\begin{figure}[H]
\centering
\includegraphics[width=1.0\textwidth]{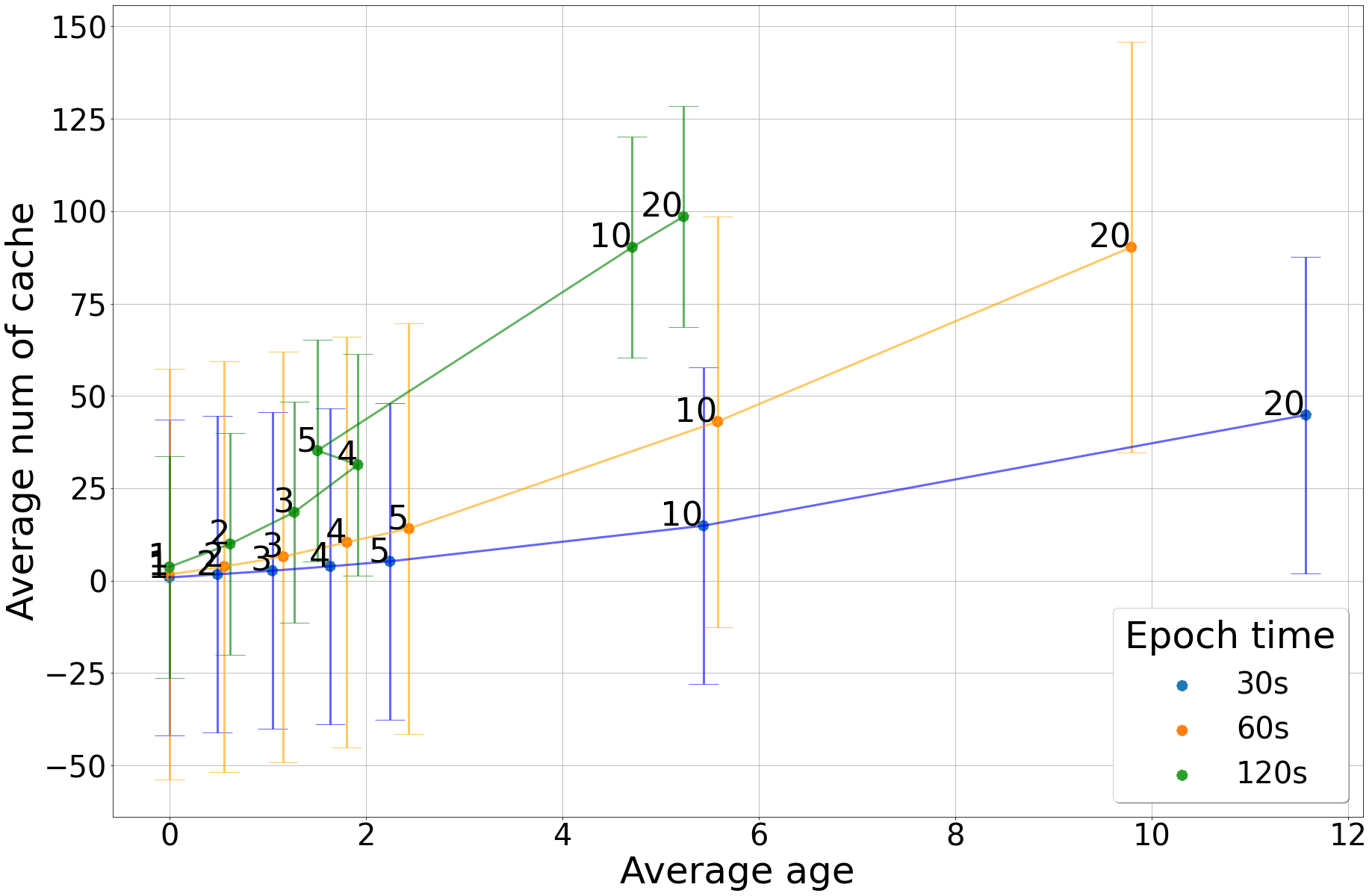}
\caption{\label{fig:cache_size_age_tradeoff_var_age} Relation Between Average Number and Average Age of Cached Models with Variation of Average Age.}
\end{figure}
\begin{figure}[H]
\centering
\includegraphics[width=1.0\textwidth]{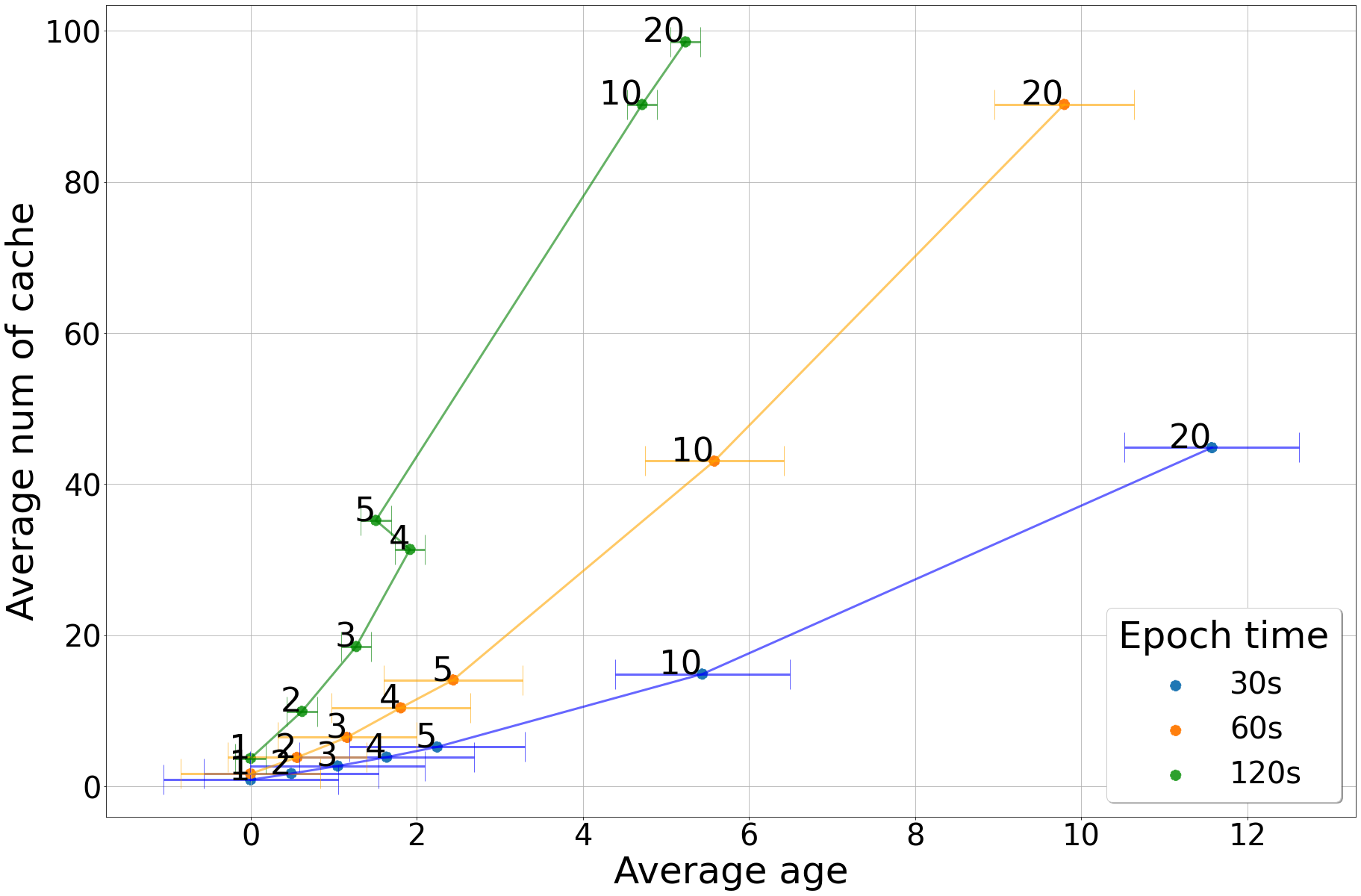}
\caption{\label{fig:cache_size_age_tradeoff_var_num} Relation Between Average Number and Average Age of Cached Models with Variation of Average Number.}
\end{figure}
\begin{figure}[H]
\centering
\includegraphics[width=1.0\textwidth]{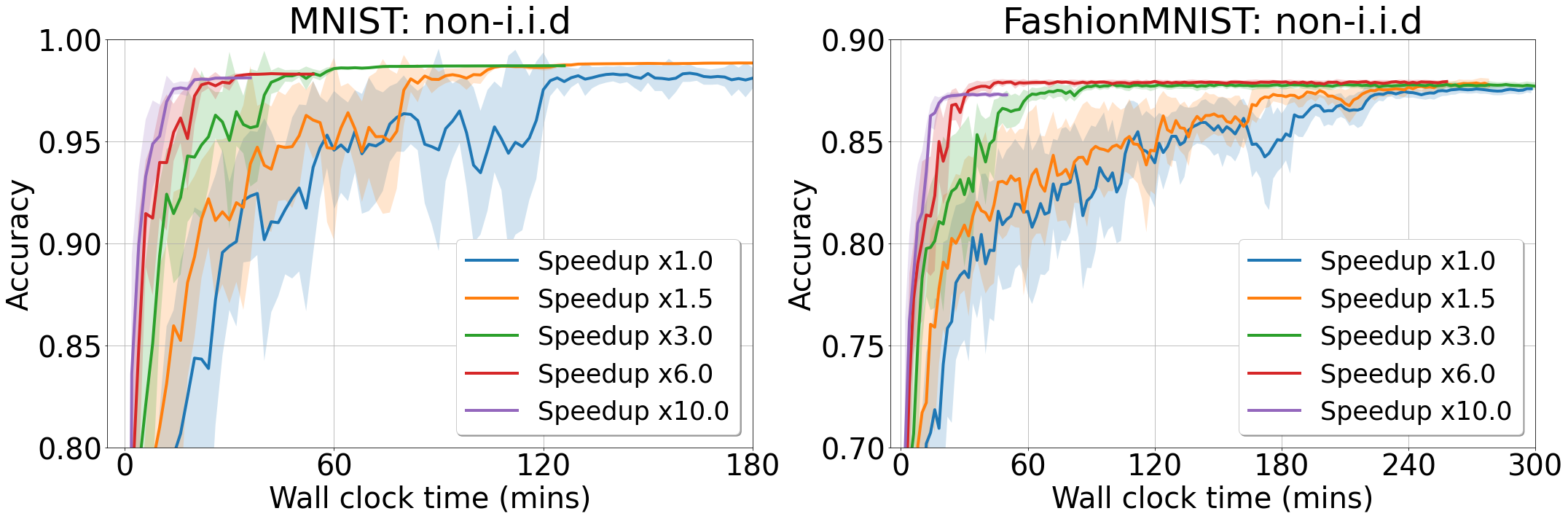}
\caption{\label{fig:speedup_var} Convergence at Different Mobility Speed with Variation.}
\end{figure}

\begin{figure}[H]
\centering
\includegraphics[width=1.0\textwidth]{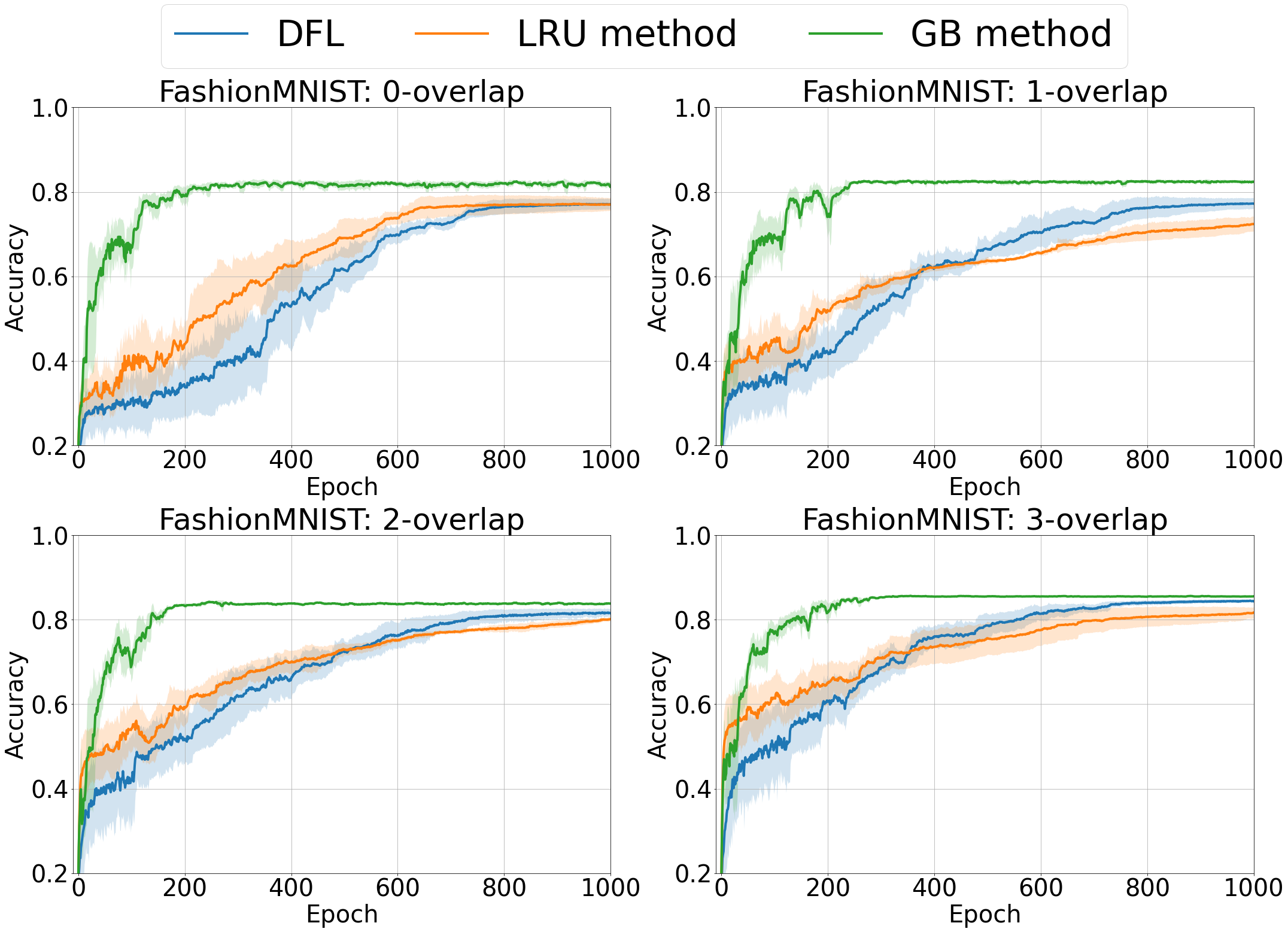}
\caption{\label{fig:Fashionmnist_GB_update_var} Group-based LRU Cache Update Performance under Different Data Distribution Overlaps  with Variation.}
\end{figure}

\begin{figure}[H]
\centering
\includegraphics[width=1.0\textwidth]{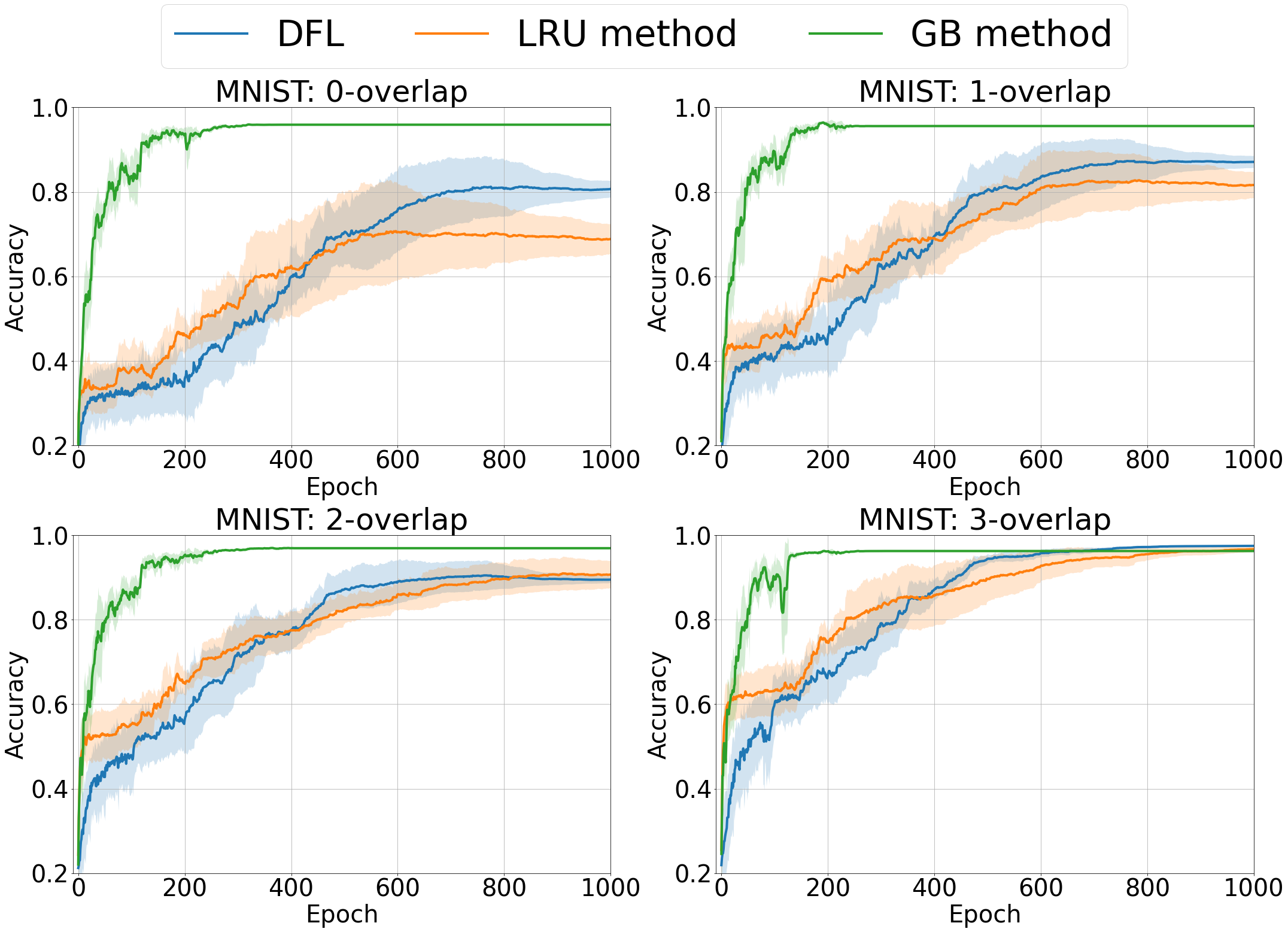}
\caption{\label{fig:mnist_GB_update_var} Group-based LRU Cache Update Performance under Different Data Distribution Overlaps  with Variation.}
\end{figure}

\subsection{Final (hyper-)parameters in experiments}
learning rate $\eta=0.1$, batch size = 64. For all the experiments, we train for 1,000 global epochs, and implement early stop when the average test accuracy stops increasing for at least 20 epochs. Without specific explanation, we take the following parameters as the final default parameters in our experiments. Local epoch $K=10$, vehicle speed $v=13.59m/s$, communication distance $100m$, communication interval between each epoch $120s$, $\tau_{max}=10$, number of devices $N=100$, cache size $10$.
\subsection{NN models}
We choose the models from the classical repository \footnote{ Federated-Learning-PyTorch
, Implementation of Communication-Efficient Learning of Deep Networks from Decentralized Data © 2024 GitHub, Inc.\url{https://github.com/AshwinRJ/Federated-Learning-PyTorch}.}: we run convolutional neural network (CNN) as shown in Table \ref{tab:CNN_MNIST}, for MNIST, convolutional neural network (CNN) as shown in Table \ref{tab:CNN_FashionMNIST}, for FashionMNIST. Inspired by \cite{zhang2019lookahead}, we choose ResNet-18 for Cifar-10, as shown in Table \ref{tab:ResNet-18}.  
\begin{table}[H]
\centering
  \caption{CNN Architecture for MNIST}
  \label{tab:CNN_MNIST}
  \begin{tabular}{cc}
    \hline
     \textbf{Layer Type} & \textbf{Size} \\ \hline
     Convolution + ReLU&$5\times5\times10$\\
     Max Pooling&$2\times2$\\
     Convolution + ReLU + Dropout&$5\times5\times20$\\
     Dropout(2D)& $p=0.5$\\
     Max Pooling&$2\times2$\\
     Fully Connected + ReLU& $320\times50$\\
     Dropout & $p=0.5$\\
     Fully Connected &$50\times 10$\\
\hline
\end{tabular}
\end{table}

\begin{table}[h]
\centering
\caption{CNN Architecture for FashionMNIST}
\label{tab:CNN_FashionMNIST}
\begin{tabular}{cc}
\hline
\textbf{Layer Type} & \textbf{Size} \\ \hline
Convolution + BatchNorm + ReLU  & $5 \times 5 \times 16$ \\ 
Max Pooling         & $2 \times 2$ \\ 
Convolution + BatchNorm + ReLU  & $5 \times 5 \times 32$ \\ 
Max Pooling         & $2 \times 2$ \\ 
Fully Connected     & $7 \times 7 \times 32\times10$ \\ \hline
\end{tabular}
\end{table}

\begin{table}[h]
\centering
\caption{ResNet-18 Architecture}
\label{tab:ResNet-18}
\begin{tabular}{cc}
\hline
\textbf{Layer Type} & \textbf{Output Size} \\ \hline
Convolution + BatchNorm + ReLU & $32 \times 32 \times 64$ \\ 
Residual Block 1 (2x BasicBlock) & $32 \times 32 \times 64$ \\
Residual Block 2 (2x BasicBlock) & $16 \times 16 \times 128$ \\ 
Residual Block 3 (2x BasicBlock) & $8 \times 8 \times 256$ \\ 
Residual Block 4 (2x BasicBlock) & $4 \times 4 \times 512$ \\ 
Average Pooling & $1 \times 1 \times 512$ \\ 
Fully Connected & $1 \times 1 \times 10$ \\ \hline
\end{tabular}
\end{table}

\end{document}